\documentclass[10pt,twocolumn,letterpaper]{article}
\usepackage{cvpr} % CAMERA-READY version
\usepackage{times}
\usepackage{epsfig}
\usepackage{graphicx}
\usepackage{amsmath,amsfonts,amssymb,mathrsfs}
\usepackage{color,soul}
\usepackage[svgnames,table]{xcolor}
\usepackage{epsfig}
\usepackage{makecell}
\usepackage{enumitem}
\usepackage{tabularx,ragged2e}
\usepackage{booktabs}
\usepackage[hang,marginal]{footmisc} % No indentation of footnotes
\usepackage{mwe}
\usepackage{graphbox}
\usepackage{caption}
\usepackage{subcaption}
\usepackage[pagebackref=true,breaklinks=true,letterpaper=true,colorlinks,bookmarks=false]{hyperref}
\usepackage[accsupp]{axessibility}

\def\cvprPaperID{2533}
\def\confName{CVPR}
\def\confYear{2022}

\usepackage{listings}
\usepackage{color}

\newcommand{\bh}{{\boldsymbol{h}}}

\newcommand{\bx}{{\boldsymbol{x}}}
\newcommand{\by}{{\boldsymbol{y}}}

\newcommand{\bT}{{\boldsymbol{T}}}

\newcommand{\btheta}{{\boldsymbol{\theta}}}
\newcommand{\bphi}{{\boldsymbol{\phi}}}

\newcommand{\pp}[1]{{\tiny{$\pm${#1}}}}
\newcommand{\hangBoxC}[1]{%
  \begin{minipage}[c]{\textwidth}\begin{tabbing} % tabbing so that minipage shrinks to fit
  ~\\[-\baselineskip] % Make first line zero-height
  #1 % Include user's text
  \end{tabbing}%^
  \end{minipage}}

\newcommand{\myparagraphWoSpacing}[1]{\vspace{-2pt}\noindent{\normalsize\bfseries #1}\hspace{.05em}} % Bold paragraph titles
\newcommand{\myparagraph}[1]{\vspace{4pt}\noindent{\normalsize\bfseries #1}\hspace{.05em}} % Bold paragraph titles

\definecolor{myBlue}{rgb}{0.1,0.1,0.8}
\definecolor{myGreen}{rgb}{0.1,0.8,0.1}
\definecolor{myRed}{rgb}{0.65,0.0,0.0}

\begin{document}

\title{Evading the Simplicity Bias:\\Training a Diverse Set of Models Discovers\\Solutions with Superior OOD Generalization}

\author{%
	Damien Teney$^{1,2}$ \quad Ehsan Abbasnejad$^2$ \quad Simon Lucey$^2$ \quad Anton van den Hengel$^{2,3}$\vspace{2pt}\\
	$^1$Idiap Research Institute~
	$^2$Australian Institute for Machine Learning, University of Adelaide~
	$^3$Amazon\\
	{\small\texttt{damien.teney@idiap.ch,\{ehsan.abbasnejad,simon.lucey,anton.vandenhengel\}@adelaide.edu.au}}\\
}

\maketitle

\begin{abstract}
Neural networks trained with SGD were recently shown to rely preferentially on linearly-predictive features and can ignore complex, equally-predictive ones.
This simplicity bias can explain their lack of robustness out of distribution (OOD).
The more complex the task to learn, the more likely it is that statistical artifacts (i.e. selection biases, spurious correlations) are simpler than the mechanisms to learn.

\vspace{5pt}
\noindent
We demonstrate that the simplicity bias can be mitigated and OOD generalization improved.
We train a set of similar models to fit the data in different ways 
using a penalty on the alignment of their input gradients.
We show theoretically and empirically that this induces the learning of more complex predictive patterns.

\vspace{5pt}
\noindent
OOD generalization fundamentally requires information beyond i.i.d. examples, such as multiple training environments, counterfactual examples, or other side information.
Our approach shows that we can defer this requirement to an independent model selection stage. %, allowing more flexibility.
We obtain SOTA results in visual recognition on biased data and generalization across visual domains.
The method --~the first to evade the simplicity bias~-- highlights the need for a better understanding and control of inductive biases in deep learning.

\vspace{5pt}
\noindent
\textbf{Addendum (09-2022)}: we subsequently used the method in~\cite{teney2022id} and extended it in~\cite{teney2022predicting} to require fewer models.
\end{abstract}
\vspace{-10pt}

%====================================================================================================================
\section{Introduction}
\label{secIntro}

\begin{figure}[t!]
	\centering
  \begin{subfigure}[b]{1\linewidth}
    \centering
    {\hspace{3.55em} \footnotesize \textbf{Training set\hspace{9.5em}OOD Test set}}\vspace{2pt}\\
    \includegraphics[width=1\linewidth]{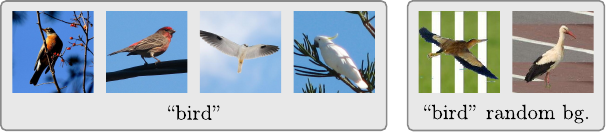}\vspace{-1pt}
    \caption{\footnotesize ImageNet-9~\cite{xiao2020noise}\label{figTeaser1}}\vspace{5pt}
  \end{subfigure}
  \begin{subfigure}[b]{1\linewidth}
    \centering
    \includegraphics[width=1\linewidth]{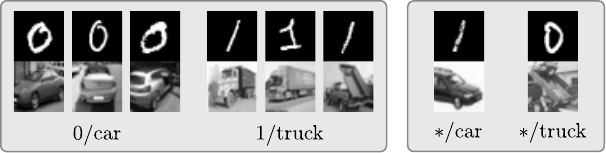}\vspace{-1pt}
    \caption{\footnotesize MNIST/CIFAR Collages~\cite{shah2020pitfalls}\label{figTeaser2}}\vspace{0pt}
  \end{subfigure}
	\caption{%
  Training data often contains multiple predictive patterns.
  %For example,
  (a)~In ImageNet-9, bird shapes and blue backgrounds are both predictive of the \textit{bird} label.
  (b)~In MNIST/CIFAR collages, both parts are equally predictive of training labels.
  Neural networks display a \textbf{simplicity bias}: they latch on the MNIST digit for example, and completely ignore the CIFAR part.
  We show that we can mitigate the simplicity bias by training a collection of models, allowing us to discover a diverse set of predictive patterns.\label{figTeaser}}
	\vspace{-2pt}
\end{figure}

\myparagraphWoSpacing{Inductive biases in deep learning.}
At the core of every learning algorithm are a set of inductive biases~\cite{mitchell1980need}.
They define the learned function outside of training examples and they allow extrapolation%
\footnote{Contrary to popular belief, deep neural networks rarely perform interpolation even in i.i.d. settings. In high dimensions, test points are extremely unlikely to lie within the convex hull of training points~\cite{hastie2009elements,krueger2020out}.}
to novel test points.
Deep neural networks are remarkably effective because their inductive biases happen to reflect properties of real-world data, although the reasons are still poorly understood~\cite{zhang2016understanding}.
In particular, the \emph{simplicity bias}~\cite{hermann2020shapes,nakkiran2019sgd,neyshabur2014search,pezeshki2020gradient,shah2020pitfalls} has been proposed as a reason of their success.
It makes networks trained with SGD%
\footnote{The simplicity bias is not a property of neural networks themselves, but also of their training with SGD, since it is possible to manually construct networks with arbitrarily poor generalization~\cite{zhang2016understanding} \ie no simplicity bias.}
 represent preferentially simple, approximately piecewise linear functions.%
\footnote{We adopt the definition of \emph{simplicity} of a feature from~\cite{shah2020pitfalls}: it is the minimum number of linear pieces in the decision boundary that achieves optimal classification accuracy using this feature. The definition naturally extends to the simplicity of a \emph{model} implementing this decision boundary.}
But the simplicity bias can also prevent the learning of complex patterns that are the actual mechanisms of the task of interest.
This effect is problematic when the learned simple patterns correspond to spurious correlations a.k.a. statistical shortcuts~\cite{geirhos2020shortcut}.
In image recognition, an example of a shortcut is to use the background rather than the shape of the object. In natural language understanding, an example is to use the presence of certain words rather than the overall meaning of a sentence.
These shortcuts are inevitable byproducts the data collection process \eg from selection biases. They are increasingly problematic as tasks tackled with deep learning grow in complexity.
The mechanisms to learn are more and more likely to be overshadowed by simpler spurious patterns.

\myparagraph{Role of inductive biases in OOD generalization.}
OOD Generalization or strong generalization is the capability of making accurate predictions under arbitrary covariate shifts.
To achieve this, a model must learn and reflect the intrinsic (\ie causal) mechanisms of the task of interest.%
\footnote{OOD Generalization goes beyond the in-domain (ID) generalization of classical learning theory. Perfect ID generalization (\ie reaching Bayes error rate on a test set from the same distribution as the training data) is achievable with infinite training data, but the predictions may rely entirely on spurious correlations (\eg recognizing birds from blue skies).}
For example, recognizing objects in arbitrary scenes requires a model to learn about their shape and details. It cannot rely solely on the background or contextual cues (Figure~\ref{figTeaser1}) .
OOD Generalization fundamentally requires extra information beyond i.i.d. training examples~\cite{bareinboim2020pearl,scholkopf2021toward}.
Existing methods use side information such as multiple training environments~\cite{arjovsky2019invariant,teney2020unshuffling,peters2016causal}, counterfactual examples~\cite{heinze2017conditional,teney2020learning}, or non-stationary time series~\cite{halva2020hidden,hyvarinen2017nonlinear,pfister2019invariant}.
Importantly, OOD generalization is not achievable only through regularizers, network architectures, or unsupervised control of inductive biases~\cite{bareinboim2020pearl}.
To make this limitation intuitive, consider the task of image recognition in Figure~\ref{figTeaser}.
Should a \emph{bird} label result from a bird shape or from a blue sky?
%One needs to determine which features are \emph{causally} related to training labels, rather than simply correlated with them because of selection biases or annotation artefacts.
If shape and background are equally predictive of training labels, the data simply lacks the information to prefer one over the other (\ie the task is underspecified: the same data could support a task where the labels relate to the background and not the objects).
This is where a learning algorithm's inductive biases come into play, possibly detrimentally.
The simplicity bias favors the most linearly-predictive patterns, but these may be spurious.
While existing methods attempt to integrate extra information during training, we show this can be deferred to a model selection stage.

\myparagraph{This study.}
We seek to control the simplicity bias of neural networks and investigate benefits in OOD generalization.
Variations of architectures, hyperparameters, or random seeds have no effect on the simplicity bias.
Instead, we train a collection of similar models to fit the training data in different ways.
Each model is optimized for standard empirical risk minimization (ERM) while a regularizer encourages diversity across the collection.
It pushes each model to rely on different patterns in the data, including complex ones that are otherwise ignore because of the simplicity bias.
Identifying a model with good OOD performance is reduced to an independent model selection step that can use any type of side information such as those mentioned above.%
\footnote{Model selection for OOD performance cannot be achieved with a standard (in-domain, ID) validation set~\cite{bareinboim2020pearl}: high ID performance can be attained by relying on spurious patterns, which says nothing about the model's capabilities OOD. An OOD validation set is a valid option that makes this step similar to the cross-validation routinely used to select architectures and hyperparameters.}

%In particular, a standard in-domain (ID) validation set is \textbf{not sufficient}~\cite{bareinboim2020pearl} (high ID performance can be attained by relying on spurious patterns, which says nothing about the model's capabilities OOD).
%An OOD validation set however is acceptable, and performing model selection by cross-validation makes this step similar to the selection routinely performed to select architectures and hyperparameters.
%Our two-step approach also allows other options, for example contrast examples~\cite{gardner2020evaluating} or model inspection with expert knowledge of features being relevant~\cite{ribeiro2020beyond,selvaraju2017grad}.

\myparagraph{Applicability of our method.}
We use three image recognition datasets to demonstrate improvements in generalization relevant to computer vision.
Issues with OOD generalization are also root causes of adversarial vulnerabilities~\cite{hendrycks2019benchmarking}, some model biases~\cite{nanda2020fairness,sagawa2020investigation}, and poor cross-domain/-dataset transfer~\cite{torralba2011unbiased}. % to name a few (see Appendix~\ref{appendix-relatedWork} for a literature review).
Potential benefits in these areas remain to be investigated. Kariyappa \etal~\cite{kariyappa2019improving} already demonstrated improved adversarial robustness by increasing diversity in an ensemble with a method similar to ours.

\myparagraph{Summary of contributions.}
\setlist{nolistsep,leftmargin=*}
\begin{enumerate}[topsep=1pt,itemsep=2pt]
  \item We review the fundamental requirements for OOD generalization and derive a rationale for addressing generalization during model selection rather than training.
  \item We describe a method to overcome the simplicity bias by learning a collection of diverse predictors.
  \item We demonstrate these benefits on existing benchmarks.
  \begin{enumerate}[topsep=1pt,itemsep=1pt]
    \item A new capability to learn multiple predictive patterns otherwise ignored because of the simplicity bias (multi-dataset collages~\cite{shah2020pitfalls}).
    \item Improved activity recognition after training on visually-biased data (Biased Activity Recognition dataset~\cite{nam2020learning}).
    \item Improved object recognition across visual domains (PACS dataset~\cite{li2017deeper}).
  \end{enumerate}
\end{enumerate}
Far from a complete solution to OOD generalization, this paper highlights the need for a better understanding and control of inductive biases in deep learning.
%Code and data are available at \url{http://<anonymized>}.

%====================================================================================================================
\section{Background}
\label{secBackground}

\myparagraphWoSpacing{Simplicity bias.}
Deep learning is actively studied to understand reasons for its successes and failures.
The simplicity bias~\cite{neyshabur2014search,shah2020pitfalls},
gradient starvation~\cite{pezeshki2020gradient}, 
and the learning of functions of increasing complexity~\cite{pezeshki2020gradient}
help explain the lack of robustness of deep neural networks and why their performance degrade under minor distribution shifts and adversarial perturbations.
Shah \etal.~\cite{shah2020pitfalls} showed that neural networks trained with SGD are biased to learn the simplest predictive features in the data while ignoring others.
Worryingly, approaches like ensembles and adversarial training --~believed to improve generalization and robustness~-- are ineffective at mitigating the simplicity bias.

\myparagraph{Shortcut learning}
~\cite{geirhos2020shortcut,lapuschkin2019unmasking} is synonymous with poor OOD generalization.
It happens when a model learns predictive patterns that do not correspond to the task of interest.
For example in object recognition (Figure~\ref{figTeaser}), the model uses the background rather than the shape of an object~\cite{baker2018deep,geirhos2018imagenet}.
The model is accurate on in-domain (ID) data (\ie from the same distribution as the training set) but is correct for the wrong reasons.
Failures are apparent on OOD test data where the spurious patterns learned during training cannot be relied on.
The more complex the task, the larger the space of possible spurious patterns that are simpler than the mechanisms of task, and the more likely is a case of shortcut learning.
The simplicity bias exacerbates shortcut learning.

%Annotation artifacts in Natural Language Inference data
%non-catastrophical shortcut learning: 2020 CrossTransformers - Spatially-aware few-shot transfer

\myparagraph{OOD Generalization}
(\ie avoiding shortcut learning) is fundamentally not attainable solely with ERM~\cite{scholkopf2021toward}. 
ERM learns \emph{any} pattern predictive of training labels.
OOD generalization requires knowing which patterns correspond to causal mechanisms of the task~(Figure~\ref{figTeaser}).
This information is lost by sampling i.i.d. training examples from the joint distribution produced by the data-generating process~\cite{bareinboim2020pearl}.
Current approaches to recover the missing information use multiple training environments~\cite{arjovsky2019invariant,teney2020unshuffling,peters2016causal}, counterfactual examples~\cite{heinze2017conditional,teney2020learning}, or non-stationary time series~\cite{halva2020hidden,hyvarinen2017nonlinear,pfister2019invariant}.
Other options to improve OOD generalization rely on ad hoc task-specific knowledge~\cite{bahng2020learning,clark2020learning,mahabadi2019simple,nam2020learning,stacey2020avoiding,utama2020mind}.

\myparagraph{Ensembles.}
This paper is not about building ensembles.
Ensembling means that multiple models are combined for inference.
Rather, we train a collection of models and identify one for inference (experiments include ensembles for comparison).
The goal of ensembling is to combine models with uncorrelated errors into one of lower variance. %(\eg decision trees combined into a random forest).
Our goal is to discover predictive patterns normally missed by a learning algorithm because of its inductive biases.

\vspace{3pt}
\noindent
See Appendix~\ref{appendix-relatedWork} for an extended literature review.

%====================================================================================================================
\vspace{-3pt}
\section{Proposed method}

\begin{figure}[t!]
	\centering
	\includegraphics[width=1\linewidth]{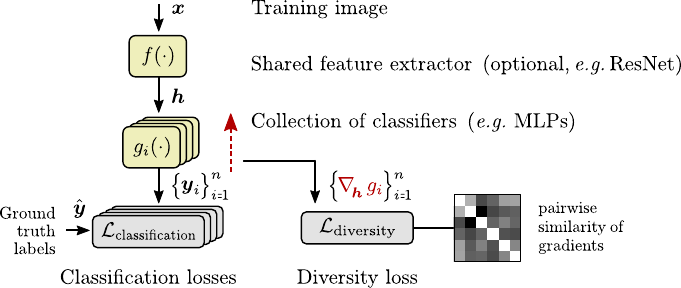}
	\vspace{-14pt}
	\caption{Our method trains a collection of classifiers in parallel to produce different predictions on OOD data.
  A diversity loss penalizes pairwise similarities between models, using each classifier's \textcolor{myRed}{input gradient} at training points.
  Combined with standard classification losses, we optimize models for both distinctness and predictive performance.\label{figMethod}}
	\vspace{-4pt}
\end{figure}

\myparagraphWoSpacing{Method overview.}
We train a collection of models in parallel (see Figure~\ref{figMethod}).
A \emph{diversity regularizer} encourages them to represent different functions.
They share the same architecture and data.
The regularizer is required because trivial options such as training models with different initial weights, hyperparameters, architectures, or shuffling of the data
do not prevent converging to very similar solutions affected by the simplicity bias as demonstrated in~\cite{shah2020pitfalls}.

%====================================================================================================================
\myparagraph{Setup.}
We consider a supervised learning task, where one model is typically trained on a training set of examples
${\bT}{=}\{(\bx^{k},\hat{\by}^{k})\}_{{k}{=}1}^{K}$. The vectors $\bx$ represent input data such as images and $\by$, in the case of a classification task, vectors of ground truth scores $[0,1]^C$ over $C$ classes.
The standard practice is to train a model (typically a neural network) for empirical risk minimization (ERM) on $\bT$.
A model implements a function $F:\operatorname{supp}(\bx) \rightarrow \operatorname{supp}(\by)$.
We represent it as a composition $F = g  \,{\circ}\,  f$ of a feature extractor $f_\btheta(\cdot)$ and classifier $g_\bphi(\cdot)$ parametrized by weights $\btheta$ and $\bphi$ respectively.
We further define the hidden representation $\bh\,{=}\,f_\btheta(\bx)$.
The model $F$ is typically optimized to minimize the risk $\mathcal{R}$ of a predictive loss $\mathcal{L}_\text{classification}$ on $\bT$ by solving
\abovedisplayskip=5pt
\belowdisplayskip=5pt
\begin{align}
  \min_{\btheta,\bphi} \; &\mathcal{R}(F_{\btheta,\bphi}) \label{eqOpti} \\[-1pt]
  \textrm{with the risk}~~ &\mathcal{R}(F_{\btheta,\bphi}) \,=\; \Sigma_{k}^K \, \mathcal{L}_\text{classification}(\hat{\by}^{k}, \by^{k}) \\[2pt]
  \textrm{and predictions}~~ &\by^{k} \,=\, F_{\btheta,\bphi}(\bx^{k}) \,=\, g_\bphi\big(f_\btheta(\bx^{k})\big) .
\end{align}
In the following, we call a \emph{predictor} any function $F_{\btheta^\star,\bphi^\star}$ from the chosen hypothesis space (\eg neural networks of a certain architecture) where $(\btheta^\star,\bphi^\star)$ is a solution to~(\ref{eqOpti}).

%====================================================================================================================
\myparagraph{Why we sometimes need more complexity.}
The simplicity bias exposed in~\cite{shah2020pitfalls} implies that a neural network trained with SGD for~(\ref{eqOpti}) relies on the simplest features predictive of labels in $\bT$.
It ignores more complex ones even if equally predictive.
The simplicity of a feature is defined in~\cite{shah2020pitfalls} as the minimum number of linear pieces in the decision boundary achieving optimal classification using this feature.
We further assume that a predictor relying this feature implements its corresponding simple decision boundary (although not stated explicitly in~\cite{shah2020pitfalls} it seems supported by their experiments).
If a simple spurious pattern exists in the data, the simplicity bias will prevent from learning any more complex mechanism of the task. 
In such cases, it is desirable to force learning a more complex predictor.

%====================================================================================================================
\myparagraph{How diversity can induce complexity.}
\label{secDiversityInducesComplexity}
By assumption of the simplicity bias, the default predictor learned by solving (\ref{eqOpti}) with SGD is the simplest.
In other words, the model learned by default lies at one end of the space of solutions.
A diverse set of solutions departing from the default one will necessarily include more complex models, that represent more complex decision boundaries and rely on different features of the data.

%====================================================================================================================
\myparagraph{How to quantify diversity.}
We compare the functions implemented classifiers using their input gradients \ie the gradient of their output with respect to their input.
Given any two classifiers $g_{\bphi_1}$ and $g_{\bphi_2}$ we quantify their similarity at a point $\bh \in \operatorname{preimage}(g)$ with
\abovedisplayskip=6pt
\belowdisplayskip=5pt
\begin{equation}
  \delta_{g_{\bphi_1},g_{\bphi_2}}(\bh) ~=~ \nabla_\bh \, g^\star_{\bphi_1}(\bh) \;.\; \nabla_\bh \, g^\star_{\bphi_2}(\bh)
  \label{eqDotProd}
\end{equation}
where the dot product measures the alignment of gradients.
Since $g$ is vector-valued, we denote with $\nabla g^\star$ the gradient of its largest component (top predicted score).
We apply (\ref{eqDotProd}) below to encourage diversity over a collection of models.

%====================================================================================================================
\myparagraph{Complete proposed method.}
Instead of training one model, we train a collection of models $\{F_i\}$  in parallel, where $F_i = g_{\bphi_i} \,{\circ}\, f$.
They share an optional feature extractor $f$ (\eg a ResNet) for computational reasons,
whereas the model-specific classifiers $g_{\bphi_i}$ are small multi-layer perceptrons (MLPs) in our experiments.
We replace the training objective~(\ref{eqOpti}) with
\abovedisplayskip=9pt
\belowdisplayskip=5pt
\begin{equation}
  \min_{(\btheta,\{\bphi_i\})} ~ \Sigma_i^n \; \mathcal{R}(g_{\bphi_i} \circ f_\btheta) ~+~ \lambda \; \Sigma_{i \neq j} \Sigma_{k}^K \; \delta_{g_{\bphi i},g_{\bphi j}}(\bh^{k}) \label{eqOpti2}
\end{equation}
where the scalar $\lambda$ controls the strength of the regularizer.
The first term is the ERM objective. It ensures a low training error and usual asymptotic guarantees for in-domain data.
The second term is the diversity regularizer. It minimizes the alignment of input gradients over pairs of models at all training points.
We solve (\ref{eqOpti2}) by SGD, with $\{\btheta_i\}$ initialized differently to break the initial symmetry.

%====================================================================================================================
\subsection{Additional considerations}
\label{secTheory}

\myparagraphWoSpacing{Rationale for input gradients.}
Intuitively, we want each model to rely on different features in the data. And input gradients are indicative of the features used by the model~\cite{selvaraju2017grad}.
Input gradients have the advantage of being implementation-invariant~\cite{sundararajan2017axiomatic} (applicable to any differentiable model) and directly relevant to OOD predictions.
%Recall our goal of training each model to fit the data differently.
The predictions of a classifiers$g_{\bphi_i}$ at a test point of features $\bh$ are denoted $g_{\bphi_i}(\bh)$.
Assuming $g$ continuous and differentiable, the approximation with a first-order Taylor expansion about a nearby training point of features $\bh^\textrm{tr}$ gives
\abovedisplayskip=6pt
\belowdisplayskip=5pt
\begin{equation}
  g_{\bphi_i}(\bh) ~=~ g_{\bphi_i}(\bh^\textrm{tr}) \;+\; (\bh - \bh^\textrm{tr}) \;\nabla_\bh g^\star_{\bphi_i}(\bh^\textrm{tr}) \;.
  \label{eqTaylor}
\end{equation}
The ERM objective clamps the value at training points which makes the first term identical $\forall \, i$.
But the diversity regularizer makes the second term different, causing predictions to diverge more and more across models as one moves away from training points. % \ie OOD.
Simple alternatives in weight space (\eg pushing parameters apart) would not guarantee learning different functions, since two networks can be equivalent under permutations and scalings of weights.

%====================================================================================================================
\myparagraph{Evaluating the diversity of learned predictors.}
Our method produces a collection models, one of which has to be selected for inference.
The selection is necessary, just like the cross-validation routinely performed to select hyperparameters and architectures, or the ubiquitous practice of early stopping.
The choice of a selection method (see below) is orthogonal to our contribution of alleviating the simplicity bias.
Therefore, the goal of our experiments is to demonstrate an increase in diversity (in terms of OOD performance) of the models that the selection can then operate on.
Therefore we report the mean, ensemble, and maximum accuracy (oracle selection) of all models of a collection.
We also perform a cross-dataset evaluation (Section~\ref{secPacs}) to verify that the maximum accuracy is meaningful and not merely an example of ``overfitting to the test set''.

%====================================================================================================================
\myparagraph{Model selection for OOD performance.}
It is important to remember that selection for OOD performance cannot, by definition, be performed with a validation set from the same distribution as the training data~\cite{teney2020value}.
OOD generalization fundamentally requires additional information beyond i.i.d. data~\cite{bareinboim2020pearl,scholkopf2021toward} (see Section~\ref{secBackground}).
Our approach uses standard i.i.d. data during training, which means that extra information has to be brought in during model selection.
It can be as simple as a small OOD validation set, and other options include any technique used to evaluate OOD performance: contrast sets, counterfactual examples~\cite{gardner2020evaluating}, inspection through with explainability techniques and expert knowledge~\cite{ribeiro2020beyond}, etc.
This flexibility of options is possible because we only require the extra information for model selection, compared to existing OOD methods that require extra information attached to every training example
\eg in multiple training environments~\cite{arjovsky2019invariant,teney2020unshuffling,peters2016causal}, counterfactual examples~\cite{heinze2017conditional,teney2020learning}, or non-stationary time series~\cite{halva2020hidden,hyvarinen2017nonlinear,pfister2019invariant}.

%====================================================================================================================
\myparagraph{Computational cost.}
Feeding all classifiers with the same mini-batches keeps the training cost small. We found no difference with feeding different mini-batches.
Memory and compute scale linearly with the number of classifiers, but these are small MLPs with a tiny footprint compared to the shared feature extractor.
With enough memory, the operations of all models can even be parallelized, hence no decrease in throughput.
The computation of the diversity regularizer reuses the input gradients that are byproducts of the backpropagation necessary to train the classifiers.
The only added cost is in the computation of second derivatives to optimize the regularizer.
All our experiments were run on a single laptop (!) with a GeForce GTX 1050 Ti GPU.

%====================================================================================================================
%\myparagraph{Complementary theoretical support.}
%We provide in Appendix~\ref{secWhy} a complementary theoretical justification for the proposed diversity regularizer.
%TODO EHSAN.

\vspace{3pt}
\noindent
%We anticipated additional questions from readers and reviewers and address them in Appendix~\ref{secFaq}.
%We address additional technical questions in Appendix~\ref{secFaq}. % which we strongly invite reviewers to consult.
We include an FAQ with past reviews in Appendix~\ref{secFaq}.
See Appendix~\ref{secFaq} for FAQs from readers and reviewers.

%====================================================================================================================
\section{Experiments}
\label{secExp}

\noindent
We designed a set of experiments to answer two questions.
\begin{enumerate}[topsep=1pt,itemsep=1pt]
  \item Can we learn predictive patterns otherwise ignored by standard SGD and existing regularizers? (Section~\ref{secCollages}).
  \item Are these patterns relevant for OOD generalization in computer vision tasks? (Sections~\ref{secBar} and \ref{secPacs}).
\end{enumerate}
\vspace{-1pt}

%====================================================================================================================
\subsection{Multi-dataset collages}
\label{secCollages}

\begin{figure}[t!]
	\centering
  \renewcommand{\tabcolsep}{.4em}
  \renewcommand{\arraystretch}{1}
  \begin{tabular}{r c c c c}
    %~ & \multicolumn{2}{c}{\scriptsize Ordered version} & \multicolumn{2}{c}{\scriptsize Scrambled version}\vspace{-.01em}\\
    %{\begin{tabular}{r}\footnotesize Class 0\\\scriptsize Zero, pullover\vspace{-4pt}\\\scriptsize automobile, zero. \end{tabular}} & \includegraphics[align=c,width=.15\linewidth]{n4class0-4.png} & \includegraphics[align=c,width=.15\linewidth]{n4class0-3.png} & \includegraphics[align=c,width=.15\linewidth]{n4class0-2-randomOrder1.png} & \includegraphics[align=c,width=.15\linewidth]{n4class0-1-randomOrder1.png}\vspace{.8em}\\
    %{\begin{tabular}{r}\footnotesize Class 1\\\scriptsize One, coat\vspace{-4pt}\\\scriptsize truck, one.            \end{tabular}} & \includegraphics[align=c,width=.15\linewidth]{n4class1-2.png} & \includegraphics[align=c,width=.15\linewidth]{n4class1-4.png} & \includegraphics[align=c,width=.15\linewidth]{n4class1-1-randomOrder1.png} & \includegraphics[align=c,width=.15\linewidth]{n4class1-3-randomOrder1.png}\\
    {\begin{tabular}{r}\footnotesize Class 0\\\scriptsize Zero, pullover\vspace{-4pt}\\\scriptsize automobile, zero. \end{tabular}} & \includegraphics[align=c,width=.15\linewidth]{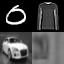} & \includegraphics[align=c,width=.15\linewidth]{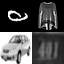} & \includegraphics[align=c,width=.15\linewidth]{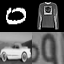} & \includegraphics[align=c,width=.15\linewidth]{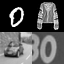}\vspace{.8em}\\
    {\begin{tabular}{r}\footnotesize Class 1\\\scriptsize One, coat\vspace{-4pt}\\\scriptsize truck, one.            \end{tabular}} & \includegraphics[align=c,width=.15\linewidth]{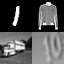} & \includegraphics[align=c,width=.15\linewidth]{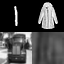} & \includegraphics[align=c,width=.15\linewidth]{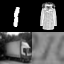} & \includegraphics[align=c,width=.15\linewidth]{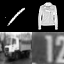}\\
  \end{tabular}
  \vspace{-5pt}
	\caption{Training examples of collages. Each block features one of two pre-selected classes from MNIST, Fashion-MNIST, CIFAR-10, SVHN.
  All four blocks are predictive of training labels. Because of the simplicity bias, a standard classifier latches on MNIST and ignores others.
  \label{figCollages}}
	%\vspace{-12pt}
\end{figure}

\myparagraphWoSpacing{Dataset.}
We extend the collages of MNIST/CIFAR (see Figure~\ref{figTeaser1}) used in previous investigations of the simplicity bias~\cite{shah2020pitfalls}.
We also use Fashion-MNIST~\cite{xiao2017fashion} and SVHN~\cite{netzer2011svhn} to form four-block collages for a binary classification task (see Figure~\ref{figCollages}).
Each block features one of two pre-selected classes from the corresponding dataset (details in Appendix~\ref{appCollagesDataset}).
In the training data, the contents of all four blocks are predictive of the labels.
Because of the simplicity bias however, a standard model systematically relies on the MNIST digit and completely ignores other parts of the image.
The dataset simulates the absence of prior preference for any image region, typical in vision tasks.
Therefore \textbf{the goal is to learn predictive patterns from all four blocks}.
This is evaluated with four test sets, in which the contents of all blocks but one are randomized to either of its two classes.
The dataset available at {\small \url{https://github.com/dteney/collages-dataset}}.

\myparagraph{Results of baselines.}
To test the simplest possible implementation of our method, we use a fully-connected 2-hidden layer MLP classifier (details in Appendix~\ref{appCollagesModels}).
We obtained upper bounds on the accuracy attainable with this architecture with four training sets where all blocks but one are randomized~(Table~\ref{tabCollagesShort}, top row).
This provides a ranking of learning difficulty: MNIST, SVHN, Fashion-MNIST, CIFAR.
We trained the baseline with popular regularizers. In all cases (32 models per experiment, repeated 5 times) the models use the MNIST digit exclusively and never perform above chance ($50$\%) on the other test sets.

\myparagraph{Results of our method.}
We then trained a collection of the same number of models (32) with our diversity regularizer.
In every case (five runs) the collection contains models that use all four parts of the images.
We determined that the models specialize but do not overlap: a model is typically good on one of the four test sets at a time (see Table~\ref{tabCollagesLong} in the appendix).
%This could be relevant for the design of a model selection strategy in practical applications.
A larger number of models also seems beneficial (see discussion in Section~\ref{secDiscussion}).
This is partly explained with the observation that, even with a large number of models, a larger fraction relies on the simpler MNIST and Fashion-MNIST blocks than on the others.
There is still room for improvement since our best models do not quite reach the upper bounds.
Finally, a manual inspection of input gradients similar to some interpretability methods~\cite{selvaraju2017grad} is an easy way to assess which part of the image is used (see Figure~\ref{figGrads}).
%with any test example 
Combined with expert task knowledge, this could serve for model selection in some applications.

\begin{table}[t]
  \footnotesize
  \renewcommand{\tabcolsep}{0.35em}
  \renewcommand{\arraystretch}{1.05}
  \centering
  \begin{tabularx}{\linewidth}{Xccccc}
  \toprule
  Collages dataset (accuracy in \%) & \multicolumn{4}{c}{Best model on} \vspace{1.5pt}\\ \cline{2-5}
  ~ &  \rotatebox{90}{MNIST} & \rotatebox{90}{SVHN~~} & \rotatebox{90}{Fashion-M.\phantom{e}} & \rotatebox{90}{CIFAR-10} & \rotatebox{90}{Average}\\
  \midrule
  Upper bounds: one predictive block in tr. & 99.7 & 89.7 & 77.4 & 68.7 & 83.9 \\
  \midrule
Baseline, 32 models with different seeds & 99.7 & 50.0 & 51.2 & 50.1 & 62.7 \vspace{-3pt}\\ ~ & \pp{0.0} & \pp{0.1} & \pp{0.3} & \pp{0.1} \\ % wt 0.00
With dropout (best rate: 0.5) & 98.7 & 54.8 & 52.9 & 54.9 & 65.3\\
%With penalty on sq. L2 norm of feature-gradient product (App.~\ref{appBaselines}) & 99.0 & 49.9 & 50.7 & 49.8 & 62.3 \\
With penalty on L1 norm of gradients & 98.9 & 49.8 & 50.7 & 49.9 & 62.3 \\
With Jacobian regularization~\cite{hoffman2019robust} & 98.8 & 49.8 & 50.7 & 49.9 & 62.3 \\
With spectral decoupling~\cite{pezeshki2020gradient} & 99.1 & 49.8 & 50.7 & 49.9 & 62.4 \\
  \midrule
%Proposed, 4 models & 98.2 & 50.1 & 52.4 & 50.8 & 62.9 \vspace{-3pt}\\ ~ & \pp{0.4} & \pp{0.2} & \pp{0.6} & \pp{0.8} \\ % wt 0.15
Proposed, 8 models & 97.3 & 82.1 & 59.6 & 55.8 & 73.7 \vspace{-3pt}\\ ~ & \pp{0.5} & \pp{6.0} & \pp{4.0} & \pp{1.9} \\ % wt 0.15
Proposed, 16 models & 96.6 & 72.1 & 64.6 & 58.4 & 72.9 \vspace{-3pt}\\ ~ & \pp{1.2} & \pp{10.3} & \pp{4.0} & \pp{1.4} \\ % wt 0.15
Proposed, 32 models & 95.6 & 81.8 & 69.2 & 61.1 & 76.9 \vspace{-3pt}\\ ~ & \pp{0.3} & \pp{5.3} & \pp{2.8} & \pp{1.0} \\ % wt 0.10
Proposed, 64 models & 95.5 & 80.9 & 70.7 & 60.8 & 77.0 \vspace{-3pt}\\ ~ & \pp{0.1} & \pp{5.8} & \pp{1.5} & \pp{0.9} \\ % wt 0.10
\textbf{Proposed}, 96 models & \textbf{95.8} & \textbf{84.7} & \textbf{71.7} & \textbf{61.7} & \textbf{78.5} \vspace{-3pt}\\ ~ & \pp{0.8} & \pp{4.0} & \pp{1.1} & \pp{1.2} \\ % wt 0.10
  \bottomrule
  \end{tabularx}
  \vspace{-6pt}
  \normalsize
  \caption{%
    Results on collages. The upper bounds are obtained by training the baseline four times on data where all blocks but one are randomized.
    Other rows correspond to the training of 32 models, of which we report the best one on each test set.
    All existing methods fixate on the MNIST block. Ours discovers predictive signals from all four blocks (mean and std. dev. over five runs).
    \label{tabCollagesShort}}
  \vspace{-12pt}
\end{table}

\begin{figure}[t!]
  \vspace{10pt}
	\centering
  \renewcommand{\tabcolsep}{.20em}
  \renewcommand{\arraystretch}{1}
  \begin{tabular}{lcccc}
    \footnotesize Baseline, 16 models~ &
    \includegraphics[align=c,width=.10\linewidth]{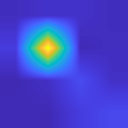}&
    \includegraphics[align=c,width=.10\linewidth]{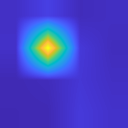}&
    \includegraphics[align=c,width=.10\linewidth]{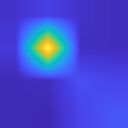}&
    \includegraphics[align=c,width=.10\linewidth]{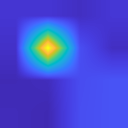}\vspace{2pt}\\
    \footnotesize \textbf{Proposed}, 16 models~ &
    \includegraphics[align=c,width=.10\linewidth]{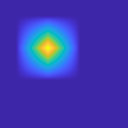}&
    \includegraphics[align=c,width=.10\linewidth]{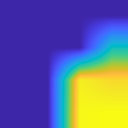}&
    \includegraphics[align=c,width=.10\linewidth]{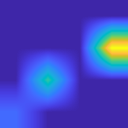}&
    \includegraphics[align=c,width=.10\linewidth]{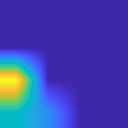}\\
    \footnotesize Best model on region: & \footnotesize ~~MNIST~~ & \footnotesize ~~SVHN~~ & \footnotesize Fashion-M. & \footnotesize ~~CIFAR~~\\
  \end{tabular} %~~\includegraphics[align=c,width=.15\linewidth]{n4class0-4.png}
  \vspace{-4pt}
	\caption{%
  Visualization of input gradients $\nabla_\bh \, g^\star(\bh)$ (abs. val. averaged over 10 test images; brighter means higher value).
  %Each model clearly specializes in a different block (quadrant).
  Each model specializes in different image regions.
  \label{figGrads}}
	%\vspace{-8pt}
\end{figure}

%====================================================================================================================
\subsection{Biased activity recognition (BAR)}
\label{secBar}

\vspace{5pt}
\begin{figure}[h!]
	\centering
  \renewcommand{\tabcolsep}{.12em}
  \renewcommand{\arraystretch}{1}
  \vspace{-12pt}
  \begin{tabular}{cccccc}
    \includegraphics[align=c,width=.16\linewidth]{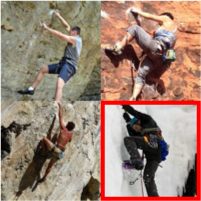} &
    \includegraphics[align=c,width=.16\linewidth]{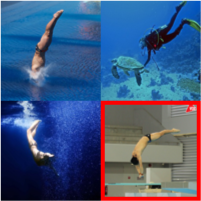} &
    \includegraphics[align=c,width=.16\linewidth]{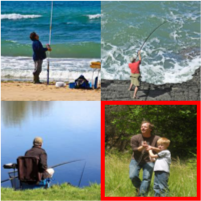} &
    \includegraphics[align=c,width=.16\linewidth]{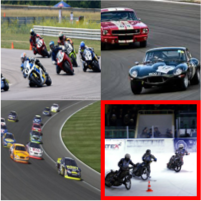} &
    \includegraphics[align=c,width=.16\linewidth]{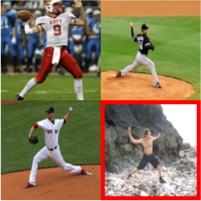} &
    \includegraphics[align=c,width=.16\linewidth]{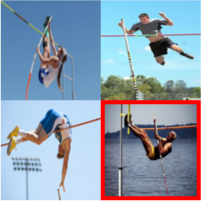} \\
    \footnotesize Climbing &
    \footnotesize Diving &
    \footnotesize Fishing &
    \footnotesize Racing &
    \footnotesize Throwing &
    \footnotesize Vaulting \\
  \end{tabular}
  \vspace{-9pt}
	%\caption{BAR dataset: three training examples, and one test example (squared in red) from each class.
  \caption{Training and test (in red) examples from BAR.
  \label{figBarExamplesSmall}}
	%\vspace{-8pt}
\end{figure}

\begin{table*}[t!]
\footnotesize
\renewcommand{\tabcolsep}{0.90em}
\renewcommand{\arraystretch}{1.06}
\centering

\begin{tabularx}{\linewidth}{Xr}
\hangBoxC{%
%\begin{tabularx}{.80\linewidth}{lccc}
\renewcommand{\tabcolsep}{0.50em}
\begin{tabular}{lccc}
\toprule
%\midrule
%\cline{2-4}
& \multicolumn{3}{c}{Biased activity recognition (BAR) dataset} \\% \cline{2-4}
\midrule
Training collection of 64 models, reporting performance of: & \textbf{Single model} & \textbf{Ensemble} & \textbf{Best single model} \vspace{-0.3pt}\\
& {\footnotesize (average accuracy} & (whole & (oracle \vspace{-1.5pt}\\
 & {\footnotesize in the collection)} & collection) & selection) \\
\midrule
Baseline in~\cite{nam2020learning} & 51.9 \pp{5.92} & N/A & N/A\\
Learning from failure~\cite{nam2020learning} & 63.0 \pp{2.76} & N/A & N/A\\
\midrule
%Our baseline, 8 models:~~~frozen ResNet-50, 2-layer MLP &
%62.2 \pp{0.4} & 63.2 \pp{0.7} & 63.4 \pp{0.4}\\
%Our baseline, 64 models:~frozen ResNet-50, 2-layer MLP &
Our strong baseline:~frozen ResNet-50, 2-layer MLP &
62.0 \pp{0.3} & 63.1 \pp{0.2} & 64.9 \pp{0.7}\\
Penalty on sq. L2 norm of gradient (Jacobian reg.~\cite{hoffman2019robust}) & 63.7 \pp{0.4} & 64.5 \pp{0.7} & 67.0 \pp{0.9}\\
Penalty on sq. L2 norm of feature-gradient product (App.~\ref{appBaselines}) & 62.8 \pp{0.1} & 63.9 \pp{0.6} & 65.9 \pp{0.5}\\
Penalty on L1 norm of feature-gradient product (App.~\ref{appBaselines}) & 63.9 \pp{0.3} & 64.6 \pp{0.4} & 66.1 \pp{0.3}\\
Penalty on sq. L2 norm of logits (spectral decoupling~\cite{pezeshki2020gradient}) & 64.3 \pp{0.2} & 65.2 \pp{0.5} & 67.0 \pp{0.4}\\
  \midrule
Proposed, 8 models &
\textbf{64.9} \pp{0.8} & 65.9 \pp{0.4} & 66.8 \pp{0.5}\\
%Proposed, 64 models:~~$\Sigma_{i{\neq}j} \, (\bh \, \nabla_\bh g_i)\miniplus . (\bh \, \nabla_\bh g_j)\miniplus$ &
\textbf{Proposed}, 64 models &
64.4 \pp{0.2} & \textbf{66.1} \pp{0.3} & \textbf{67.1} \pp{0.3}\\
%Proposed, 64 models + spectral decoupling & 64.2 \pp{0.3} & 65.2 \pp{0.3} & 66.5 \pp{0.2}\\
\bottomrule
\end{tabular}%
%\end{tabularx}%
}&\hangBoxC{\\\vspace{-10pt}
  \includegraphics[height=160pt]{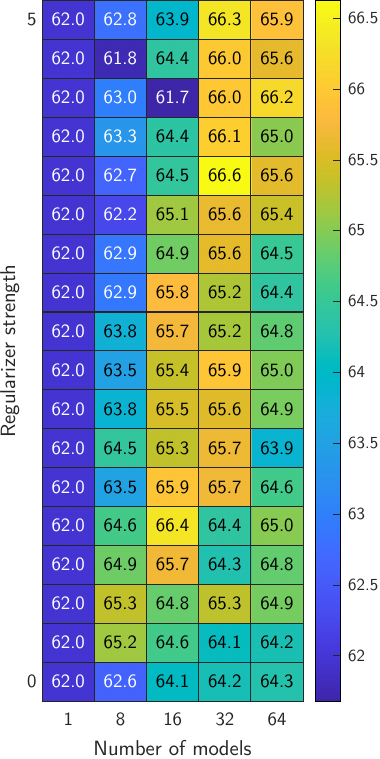}%~
  %\includegraphics[height=140pt]{barVertical3scale1.pdf}
  %\vspace{4pt}
}\\
\end{tabularx}
\vspace{-10pt}
\caption{Evaluation on BAR (mean accuracy and std. dev. over five runs). Each row corresponds to the training of 64 models, unless otherwise noted.
The heat map (right; accuracy of best model) shows that the diversity regularizer clearly improves over a classical ensemble of independently-trained models (bottom row).\label{tabBar}}
\vspace{-8pt}
\end{table*}

\myparagraphWoSpacing{Dataset.}
The BAR dataset~\cite{nam2020learning} was recently introduced to evaluate debiasing methods for image recognition.
The task is to classify photographs into six activities (see Figure~\ref{figBarExamplesSmall}).
Training images were sampled from six pairs of actions/places in imSitu~\cite{yatskar2016} \eg climbing/rock.
Test images were sampled from the same actions in different places \eg climbing/ice. % (see Figure~\ref{figBarExamples} in the appendix).
\textbf{The goal is to learn a model that relies more on a person's appearance than on the background} to recognize actions in arbitrary places.
This is challenging because both are predictive of the action training labels.
This represents an ubiquitous setting where training data is affected by selection biases but an image recognition model is still expected perform well on novel scenes.

\myparagraph{Results.}
We follow~\cite{nam2020learning} and implement a classifier on ResNet-50 features (details in Appendix~\ref{appBar}).
We first tune a strong baseline classifier (almost equating the method in~\cite{nam2020learning}).
We then train a collection of these classifiers with our regularizer.
The accuracy of the best model improves from $64.9$ to $67.1$ (Table~\ref{tabBar}).
The \emph{average} accuracy over the collection also increases, perhaps surprisingly.
Indeed, an increase in diversity could induce as many worse models than better ones.
But remember that the space of predictors ranked by complexity is one-sided (Section~\ref{secDiversityInducesComplexity}).
The baseline is at the ``simplest'' end. Those learned with our method are more complex.
With BAR, complex models happen to be better, as analyzed in~\cite{nam2020learning}.
Thanks to this, a simple ensemble (summing all predicted scores) improves over the same ensemble of models trained without our regularizer ($63.1$ $\rightarrow$ $66.1$) with no model selection.
We insist however that this is not a universal benefit of our method.

The BAR dataset contains no information to prefer backgrounds or persons' appearance. With the same training data and annotations, the task could as well be to recognize places rather than actions !
This reinforces our insistence that side information is necessary for OOD generalization. 
Debiasing methods like~\cite{nam2020learning} rely on task-specific design choices.
Our approach is of more general purpose.

The existing Jacobian regularizer~\cite{hoffman2019robust} and spectral decoupling~\cite{pezeshki2020gradient} produce models almost as good as our best one, although worse on average as seen with a lower \textit{ensemble} accuracy.
The advantage of our method is to produce a collection of diverse models, whereas any specific regularizer is either suited or not to the given task. If not, the practitioner has to manually find another one.

%====================================================================================================================
\subsection{Domain generalization (PACS)}
\label{secPacs}

\begin{table*}[t]
\footnotesize
\renewcommand{\tabcolsep}{0.10em}
\renewcommand{\arraystretch}{1.05}
\centering

\begin{tabularx}{\linewidth}{Xr}
  \hangBoxC{%
  \renewcommand{\tabcolsep}{0.25em}
  \begin{tabularx}{.76\linewidth}{Xccc c c}
    %\begin{tabular}{lccccc}
    \toprule
    Training set & \multicolumn{5}{c}{PACS (cartoon, photo, sketch)}\vspace{1.5pt}\\ \cline{2-6} %\vspace{-12pt}\\
    Test set & \multicolumn{3}{c}{PACS (art)} & ~ & VLCS (horse/person AUC)\vspace{1.5pt}\\  \cline{2-4}  \cline{6-6} %\vspace{-12pt}\\
    Model evaluated & Single & Ensemble & Best & ~ & Best model on PACS \\
    \midrule
    Baseline, 64 models, no regularizer & 84.48 \pp{0.23} & 84.62 & 85.71 & ~ & 74.57 \\
    Penalty on sq. L2 norm of grad. (Jacobian reg.~\cite{hoffman2019robust}) & 85.12 \pp{0.33} & \textbf{85.06} & 85.84 & ~ & 74.10 \\
    Penalty on sq. L2 norm of ReLU of grad. (App.~\ref{appBaselines}) & 84.62 \pp{0.19} & 84.62 & 85.16 & ~ & 75.84 \\
    Penalty on sq. L2 norm of feature-grad. prod. (App.~\ref{appBaselines}) & 84.61 \pp{0.26} & 84.77 & 85.45 & ~ & 73.51 \\
    Penalty on L1 norm of grad. (App.~\ref{appBaselines}) & 84.66 \pp{0.45} & 84.67 & 86.13 & ~ & 76.29 \\
    Penalty on sq. L2 norm of logits (spectral dec.~\cite{pezeshki2020gradient}) & 84.46 \pp{0.32} & 84.81 & 85.16 & ~ & 74.60 \\
    Combination: proposed + spectral dec.~\cite{pezeshki2020gradient} & 84.31 \pp{0.83} & 84.72 & 86.08 & ~ & 74.51 \\
    \midrule
    \textbf{Proposed}, 64 models & \textbf{85.14} \pp{0.59} & 84.62 & \textbf{86.80} & ~ & \textbf{79.66} \\
    \bottomrule
  \end{tabularx}} &
  \hangBoxC{\vspace{4pt}\\\includegraphics[height=131pt]{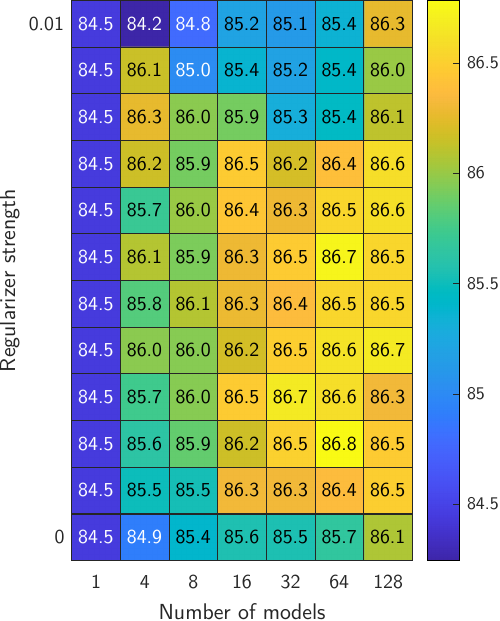}}
\end{tabularx}
\vspace{-10pt}
\caption{(Left)~Ablative evaluation on PACS (mean accuracy and std. dev. over five runs). We also evaluate models selected on PACS on test data from VLCS.
This confirms that these models do indeed generalize better.
(Right)~This heatmap (accuracy on PACS, best model) clearly shows improvements over independently-trained models (bottom row).
\label{tabPacsAblation}}
\vspace{-4pt}
\end{table*}

\begin{figure*}[t!]
	\centering
  \hangBoxC{\includegraphics[width=.52\linewidth]{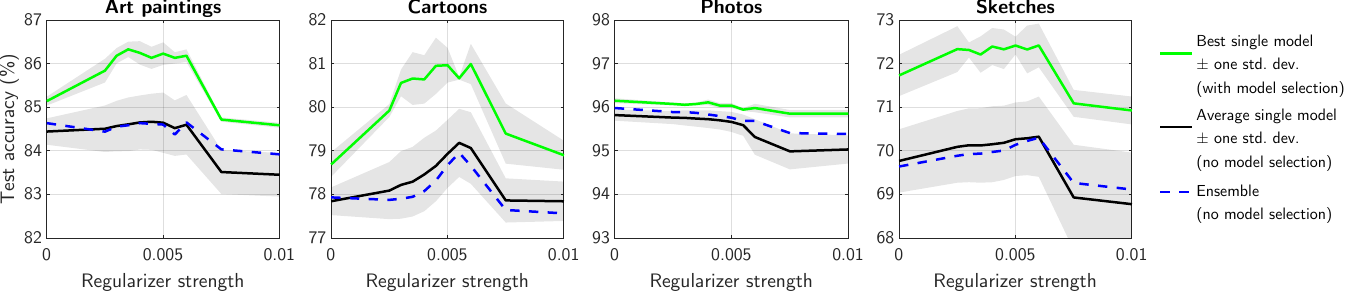}}~~~~~
  \hangBoxC{\includegraphics[width=.11\linewidth]{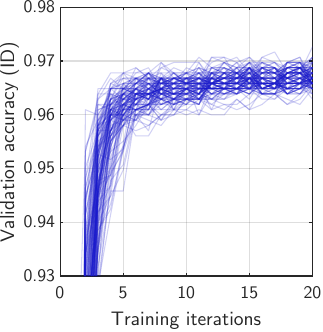}~\includegraphics[width=.11\linewidth]{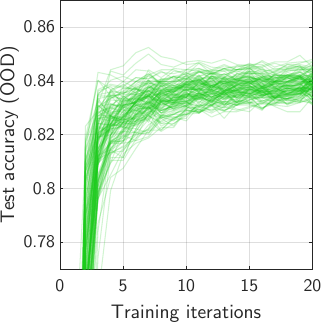}~~\includegraphics[width=.11\linewidth]{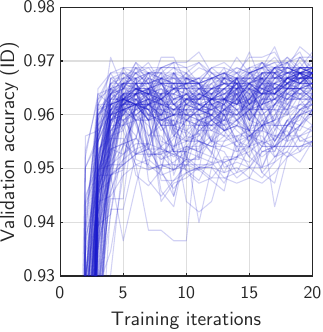}~\includegraphics[width=.11\linewidth]{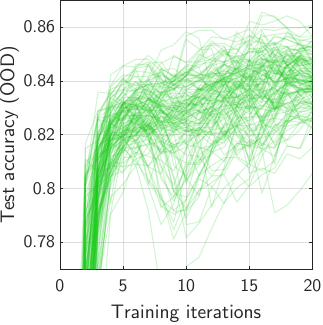}}
	\vspace{-8pt}
	\caption{(Left)~Accuracy on each test style of PACS as a function of the regularizer strength. (Right)~Training curves of \textcolor{myBlue}{in-domain} and \textcolor{myGreen}{OOD} accuracy of 64 models, either trained independently (left two plots) or with our regularizer (right twos).\label{figPacsRegWeight}}
	\vspace{-11pt}
\end{figure*}

\myparagraphWoSpacing{Dataset.}
PACS is a standard benchmark for visual domain generalization (DG)~\cite{li2017deeper}.
It contains images from seven classes and four visual domains: art paintings, cartoons, photographs, and sketches.
It is used in a leave-one-out manner with three training domains and the remaining one for testing.
The standard baseline uses images from all domains indistinctively whereas DG methods use domain labels to try and identify common features that should also be reliable in the test domain.

\myparagraph{Results.}
We report an ablative study in Table~\ref{tabPacsAblation} (see Appendix~\ref{appPacs} for implementation details).
%We report the same measures (single/ensemble/best) as described for the BAR dataset.
%As with the BAR dataset, w
We obtain a better model by training a collection with our regularizer, compared to the baseline where differences within the collection result only from different random seeds.
The improvements are modest but they clearly result from our regularizer, as seen from a \hyperref[tabPacsAblation]{heatmap of accuracy} as a function of the number of models and regularizer strength (also see Figure~\ref{figPacsRegWeight}, left).
A larger number of models also seem beneficial.
We provide the training curves of all models in a collection in Figure~\ref{figPacsRegWeight}.
Looking at the OOD accuracy as training progresses, we see that the regularizer induces vastly more variance, producing both worse and better models as expected.

As observed with BAR, the average accuracy within a collection improves slightly with the regularizer~--~although not to the point that a naive ensemble would be beneficial. This corroborates the explanations given above for BAR.

We include an additional cross-dataset evaluation \ie zero-shot transfer (Table~\ref{tabPacsAblation}, last column).
We use the test images from VLCS~\cite{ghifary2015domain} for a detection task of classes both in PACS and VLCS (\textit{horse} and \textit{person}; other VLCS classes serve as negatives).
We report the area under curve (AUC) averaged over these two classes. 
We observe that the best model \textbf{selected on PACS} also brings substantial improvements on VLCS. This verifies that the accuracy of the best model on PACS is meaningful and not merely an example of ``overfitting to the test set''.

We evaluate regularizers previously proposed to improve generalization.
We spent substantial effort tuning these alternatives to their best and trying multiple variants (see Appendix~\ref{appBaselines} for details).
% Table~\ref{tabPacsAblation} (see also Table~\ref{tabBar} for a summary of their mathematical form).
As seen in Table~\ref{tabPacsAblation}, a Jacobian regularizer provides small improvements. So does minimizing the squared L2 norm of the gradient.
Our regularizer mainly minimizes the alignment between pairs of gradients but it can also reduce their absolute norm as a side effect, just like these two alternatives.
The ablation shows this to be beneficial but also that it does not entirely explain the benefits of our method.
Indeed on VLCS, only our regularizer provides a substantial improvement ($74.57$ $\rightarrow$ $79.66$).

We provide a comparison with state-of-the-art DG methods in Table~\ref{tabPacsAllStyles}.
Most methods rely on labels of training domains, which we do not use.
Most were demonstrated on top of weak baselines, as pointed out in~\cite{gulrajani2020search}.
We use the same ResNet-18 feature extractor, we highly optimized it, and we still get substantial improvements.
See Section~\ref{secDiscussion} for more discussion about the comparison with DG methods.

\begin{table}[t]
\footnotesize
\renewcommand{\tabcolsep}{0.25em}
\renewcommand{\arraystretch}{1.01}
\centering
\begin{tabularx}{\linewidth}{Xccccc}
%\begin{tabular}{lccccc}
\toprule
%\midrule
%\cline{2-4}
%& \multicolumn{5}{c}{PACS Dataset} \\ %\cline{2-6}
%\midrule
PACS Dataset & \includegraphics[align=c,width=.081\linewidth]{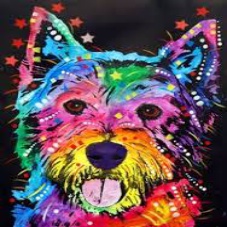} & \includegraphics[align=c,width=.081\linewidth]{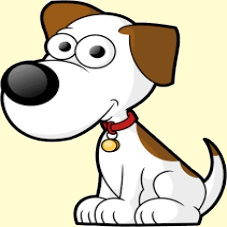} & \includegraphics[align=c,width=.081\linewidth]{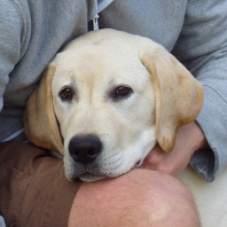} & \includegraphics[align=c,width=.081\linewidth]{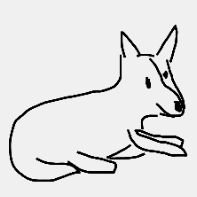} & ~\\
Test style (leave-one-out) & Art & Cartoon & Photo & Sketch & Avg.\\
\midrule
%clipboard('copy', strjoin(compose(" & %.1f", round(a,1))))
%\multicolumn{6}{l}{Methods using labels of training domains}\\
%MLDG \cite{li2018learning} & 79.5  & 77.3  & 94.3  & 71.5  & 80.7\\
%MetaReg \cite{balajii2018metareg}  & 79.5  & 75.4  & 94.3  & 72.2  & 80.4\\
%CrossGrad \cite{volpi2018generalizing} & 78.7  & 73.3  & 94.0  & 65.1  & 77.8\\
%MASF \cite{dou2019domain} & 80.3  & 77.2  & 95.0  & 71.7  & 81.0\\
%Cumix (Mancini et al. 2020) 82.30 76.50 95.10 72.60 81.60
D-SAM baseline \cite{d2018domain} & 77.9  & 75.9  & 95.2  & 69.3  & 79.6\\
D-SAM$^\ast$ & 77.3  & 72.4  & 95.3 & 77.8   & 80.7\\
\midrule
Epi-FCR baseline \cite{li2019episodic} & 77.6  & 73.9  & 94.4  & 74.3  & 79.1\\
Epi-FCR$^\ast$ & 82.1  & 77.0  & 93.9  & 73.0  & 81.5\\
\midrule
DMG baseline \cite{chattopadhyay2020learning} & 72.6  & 78.5  & 93.2  & 65.2  & 77.4\\
DMG$^\ast$ & 76.9  & 80.4  & 93.4  & 75.2  & 81.5\\
\midrule
DecAug baseline \cite{bai2020decaug} & 78.4  & 78.3  & 94.2  & 72.1  & 80.8\\
DecAug$^\ast$ & 79.0  & 79.6  & 95.3  & 75.6  & 82.4\\
\midrule
%\multicolumn{6}{l}{Other methods}\\
JiGen baseline \cite{carlucci2019domain} & 77.9  & 74.9  & 95.7  & 67.7  & 79.1\\
JiGen & 79.4  & 75.3  & 96.0  & 71.4  & 80.5\\
\midrule
Latent domains baseline \cite{matsuura2020domain} & 78.3  & 75.0  & 96.2  & 65.2  & 78.7\\
Latent domains & 81.3  & 77.2  & 96.1  & 72.3  & 81.8\\
\midrule
\multicolumn{6}{l}{Our baseline, 64 independent models}\vspace{2pt}\\

Random single model& 84.4 & 77.8 & 95.8 & 69.8  & 82.0\vspace{-3pt}\\
~& \pp{0.3} & \pp{0.3} & \pp{0.1} & \pp{0.7} & ~\\

Ensemble of all models & 84.6 & 77.9 & 96.0 & 69.6  & 82.0\vspace{-3pt}\\
~& \pp{0.1} & \pp{0.1} & \pp{0.1} & \pp{0.2} & ~\\

Best single model & 85.1 & 78.7 & \textbf{96.2} & 71.7  & 82.9\vspace{-3pt}\\
~& \pp{0.1} & \pp{0.3} & \pp{0.1} & \pp{0.5} & ~\\

\midrule
\multicolumn{6}{l}{\textbf{Proposed}, 64 models}\vspace{2pt}\\

Random single model & 85.2 & 79.6 & 95.9 & 70.8 & 82.9\vspace{-3pt}\\
~& \pp{0.6} & \pp{0.7} & \pp{0.1} & \pp{0.8} & ~\\

Ensemble of all models & 84.8 & 79.0 & 96.0 & 70.3 & 82.5\vspace{-3pt}\\
~& \pp{0.1} & \pp{0.1} & \pp{0.1} & \pp{0.1} & ~\\

Best single model & \textbf{86.5} & \textbf{81.1} & \textbf{96.2} & \textbf{72.8} & \textbf{84.2}\vspace{-3pt}\\
~& \pp{0.1} & \pp{0.4} & \pp{0.4} & \pp{0.2} & ~\\
\bottomrule
\end{tabularx}
%\vspace{-2pt}
\normalsize
\caption{%
  Comparison with existing methods on PACS.
  For each method, we mention the accuracy of the baseline reported by its authors, and of the method itself.
  All methods are based on a ResNet-18. $^\ast$Require labels of tr. domains.
  Our method discovers additional predictive features in the data. It returns a collection of models, among which some clearly have a better OOD performance (last row) than the baseline simply trained with different random seeds ($82.9 \rightarrow 84.2$).
  \label{tabPacsAllStyles}}
\vspace{-8pt}
\end{table}

%====================================================================================================================
\section{Discussion}
\label{secDiscussion}

The above results show that (1)~we now have a tool to expand the set of solutions learned by a neural network through SGD and (2)~some solutions are relevant to computer vision as evidenced by improved OOD generalization.

%====================================================================================================================
\myparagraph{Limitations of the method.}
The main hyperparameters are the regularizer strength and the number of models learned.
Any number $>$1 obviously gives more options than the single model learned by default.
Empirically, larger numbers ($>$64) seem beneficial, but we did not derive guarantees that a robust predictor will be found.
The chosen number of models may not induce the optimal ``granularity''.
Too small a number could produce a model that relies on multiple entangled features including robust and spurious ones.
And too large a number could produce multiple models that each implement a different part of a robust solution.

An interesting direction for future work would be to encourage learning a ``basis'' of elemental predictors (suggested in~\cite{ross2020ensembles}) for a given dataset by promoting notions of sparsity or complementarity.
The model selection could thereafter search for their optimal combination as attempted in~\cite{pace2020learning}.
The approach would resemble the disentanglement methods of representation learning~\cite{locatello2019challenging} but it would operate in the space of predictors optimized for ERM, rather than the space of features optimized for reconstruction.

%====================================================================================================================
\myparagraph{The need for model selection.}
The utility of evading the simplicity bias is in improving OOD performance because this is where the simplicity bias shows its detrimental effects (high ID performance is achievable with less sophistication).
The key novelty is to require no extra information \textbf{during training} contrary to existing methods that require upfront task-specific knowledge (\eg debiasing methods~\cite{bahng2020learning,clark2020learning,mahabadi2019simple,nam2020learning,stacey2020avoiding,utama2020mind}) or additional annotations~\cite{arjovsky2019invariant,peters2016causal,teney2020unshuffling,teney2020learning}.
The requirement for side information (besides a training set of i.i.d. examples) is fundamental for OOD performance~(see~\cite{bareinboim2020pearl,scholkopf2021toward} and additional background in Appendix~\ref{secCausality})
but we defer its use to an independent model selection step.
This allows more flexibility in the type and quantity of side information used.

%====================================================================================================================
\myparagraph{Limitations of the evaluation.}
In a review on domain generalization (\ie OOD generalization across visual styles), Gulrajani and Lopez-Pas~\cite{gulrajani2020search} note that model fitting and model selection are equally hard.
They recommend that methods for datasets like PACS (Section~\ref{secPacs}) include a selection strategy.
On the opposite, we show that the two steps can be completely decoupled.
Our analysis focuses on the best learned model (denoted \emph{oracle selection} in~\cite{gulrajani2020search} and \emph{best} in~\cite{pace2020learning}).
This optimistic choice is justified because it is the performance achievable with an optimal selection strategy.
\textbf{This accounts for existing and future selection strategies} \eg the calibration-based method of Wald \etal~\cite{wald2021calibration} that came out after the writing of this paper.

%Current OOD benchmarks are not ideal to evaluate this two-step approach. Idelly, they would ideally include various amounts and forms of side information to compare selection strategies.
%In our experiments, we report the performance of the best learned model since it represents the achievable performance with optimal selection.

Is the comparison unfair with methods for PACS that train a single model?
We do not think so: existing methods were selected at the \textit{paper} level. Methods with no improvement on the test set did not get published.
It is unlikely that authors never peeked at test-set performance until getting published !
It was even showed in~\cite{gulrajani2020search} that all tested methods lost their benefits when deprived of heavy hyperparameter tuning and early stopping based on OOD performance.
Our approach makes the selection more explicit.

%====================================================================================================================
\myparagraph{Heuristics about simplicity.}
Our improvements support previous claims~\cite{neyshabur2014search,shah2020pitfalls} that more complex models are sometimes preferable.
Works in NLP have even argued that any simple correlation in a dataset is likely to be spurious~\cite{clark2020learning,utama2020towards,zhang2019perturbation}.
We point out that such heuristics are necessarily task- or dataset-specific.
However, we also remark that the tasks addressed with deep learning are increasingly complex (take visual question answering for example~\cite{teney2017spm}).
This implies that the space of potentially-spurious patterns that are simpler than the task in any given dataset is also growing.
The above heuristics may therefore have a practical utility. It remains crucial to study their limits of applicability.
They cannot be universal~\cite{mitchell1980need} and cannot obviate the need for extra information (or task-specific knowledge) in our model selection step.

%====================================================================================================================
\myparagraph{Universality of inductive biases.}
The inductive biases of any learning algorithm cannot be universally superior to another's~\cite{wolpert2002supervised}.
For example, weight decay, Jacobian regularization, or even data augmentation are only as good as they are tuned to a particular task.
In comparison, our method does not affect inductive biases in a directed way. It only increases the variety of the learned models, so it could be seen as a ``meta-regularizer''.
Our results also show that intuitive notions behind classical regularizers like smoothness (Jacobian regularization), sparsity (L1 norm), or simplicity (L2 norm) are sometimes detrimental.

Any quest for \emph{universal} architectures, regularizers, data augmentations, or even dataless selection strategies~\cite{murdock2020dataless,zhang2019perturbation} is known to be futile.
Benefits can only apply to a subset of learning tasks~\cite{wolpert2002supervised}.
Questions remain: how small or vast is the subset of learning tasks that humans care about? Which properties of naturally-produced data make some inductive biases \emph{generally} useful?
Deep learning has proven surprisingly successful.
Studying its inductive biases will help understanding its limits of applicability.
And methods to control these biases will help expanding these limits.

%Future work should still include both controlled studies (\eg collages or synthetic data) and realistic challenging datasets (\eg the WILDS benchmark~\cite{koh2020wilds}).
%WILDS - Benchmark of in-the-Wild Distribution Shifts

%====================================================================================================================
% Max 8 pages + references
\clearpage
%\subsection*{Acknowledgements}
%This material is based on research sponsored by Air Force Research Laboratory and DARPA under agreement number FA8750-19-2-0501. The U.S. Government is authorized to reproduce and distribute reprints for Governmental purposes notwithstanding any copyright notation thereon.

\bibliographystyle{ieee_fullname}
\bibliography{Bibliography}

\clearpage
\appendix
\section*{Evading the Simplicity Bias:\\Training a Diverse Set of Models Discovers\\Solutions with Superior OOD Generalization}
\vspace{4pt}
\section*{Supplementary material}
\vspace{9pt}

%====================================================================================================================
\section{Past reviews and FAQ}
\label{secFaq}

We include questions and comments received during the reviewing process along with our answers.
We hope that they will be helpful to readers.

\myparagraph{Q1: Where to split a model into ``feature extractor'' and ``classifier''?}
The separation between the feature extractor and the classifier is somewhat arbitrary.
In our experiments, it was dictated by computational reasons.
Using a pretrained ResNet as the feature extractor was a natural choice.
The subsequent classifiers need sufficient width/depth such that they can model different functions on top of the features.
The choice of 2-hidden-layer MLP was a compromise between expressibility and computational cost.
A possible extension of this work would be to evaluate deeper classifiers or complete replicated models.

The choice of the ``extractor/classifier'' separation also affects the dimension along which the gradients are compared.
For example with ResNet, features undergo a global spatial pooling before being passed to the classifier.
The gradients are thus compared only across channels, not spatial dimensions.
In our experiments with collages, the feature extractor is the identity function and the classifier is a fully-connected network operating directly on the pixels. The gradients are thus compared across spatial dimensions

%We can take the gradient of the classifier's output for the max/correct/all/ classes

\myparagraph{Q2: Why not design the diversity regularizer on the activations of the models rather than on the input gradients?}
Because the activations in different models live in ``different spaces'' so they are not directly comparable.
Consider any chosen layer: the activations do not tell much about the function implemented by the whole network. Whereas the input gradients (through the whole network) do.

The activations also cannot be used with a simple dot product to compare different models.
For example, one can rearrange the weights of a model to permute the channels of the activations without actually changing the function implemented by the whole network.
%In my method, the fact that the models are implemented as neural networks (stacked layers with intermediate activations, etc.) is actually irrelevant.

\myparagraph{Q3: Is the introduction of more diversity just a fancy random search?}
The common training by SGD is already stochastic.
It is not uncommon to restart training by SGD with different random seeds if training does not converge.
But simply using different random seeds is not sufficient to evade the simplicity bias however, whereas the proposed method is.

The proposed method does not bring any additional source of randomness. It makes solutions found by SGD from different initializations more different from one another.

\myparagraph{Q4: Why not use the Colored MNIST toy data?}
A number of recent works have use ``Colored MNIST'' toy data introduced in~\cite{arjovsky2019invariant} to evaluate OOD generalization.
We believe it is an overly simplistic setting.
The so-called ``color'' only means that the MNIST pixels are stored on one discrete channel or another.
This is like operating on symbolic data, or on perfectly disentangled representations. And learning disentangled representations from high-dimensional data is itself a largely unsolved problem.
Our collages achieves a similar objective (multiple predictive signals of different complexity) but it is more challenging and representative of real data.

\myparagraph{Q5: Why not use dataset [X]?}
We chose a few datasets representative of OOD challenges in computer vision (multiple predictive signals, biased data, generalization across visual styles).
We had to choose datasets with an established OOD test set.
For the reasons explained at length in the paper, no improvement is expected with identically-distributed training and test sets.

\myparagraph{Q6: Why doesn't in-domain (ID) performance necessarily improve when OOD performance does?}
Because the irreducible error with robust features can be larger than with spurious ones.
This is not uncommon and it often makes improvements in OOD performance come at the expense of ID performance.

To do well ID, a model can exploit any predictive pattern in the training data, including spurious correlations.
For example, birds may be reliably recognized in the training data by detecting a blue background.
However, relying on blue backgrounds is not desirable for an OOD bird detector.
A good OOD model would instead rely on shape, such that it can recognize birds in any scene.
But shapes are more often ambiguous, so in ID scenes (where the blue color is a reliable indicator of birds) the OOD model will not be as accurate as a blue-background detector.

%The stage two uses out-of-distribution validation data to select a model from the collection learned in Stage one. If the distribution of the validation set and the distribution of the target test data are the same, the model selection solves a domain adaptation problem. If the distribution of the validation set and the distribution of the target test data are the different, the model selection stage cannot outperform in-distribution validation accuracy nor leave-one-domain-out validation accuracy without priori information.

\myparagraph{Q8: \textit{``In stage one, there is no guarantee that simple biases (i.e. spurious features) and complex patterns (i.e. robust features) can be disentangled. Could each model in the collection be a mixture of simple biases and complex patterns?''}}
%\vspace{3.9pt}\noindent
It is correct that no semantically or causally-meaningful disentanglement can be guaranteed.
What can be guaranteed is that the models trained in a large-enough set cannot be decomposed into ``simpler'', locally independent functions.
Indeed,~\cite{ross2018learning} showed that local independence enforced with orthogonal gradients leads to the recovery of sparse predictors. As one increases the number of predictors fitted in parallel, they approach a ``maximal set'' as defined in~\cite{ross2018learning}. This implies functional simplicity, in the sense that each predictor in a maximal set cannot be further decomposed into a combination of locally independent functions.
%(i.e. a form of disentanglement by some definitions).

\myparagraph{Q8: \textit{``Is it possible to train all 96 MLP classifiers simultaneously (in parallel) on a single GPU?''}}
Indeed we had no issue training $96$ MLPs on a laptop GPU, thanks to the sharing of mini-batches across models.

\myparagraph{Q9: \textit{``In Table 1, can you explain the decrease in accuracy when increasing the number of models from 8 to 16?''}}
We attribute this decrease in accuracy to evaluation noise since it is well within the standard deviation across runs.

\myparagraph{Q10: \textit{``Can you elaborate on the connection between this paper's findings and the simplicity bias?''}}
The connection follows from the recent evidence~\cite{shah2020pitfalls} that the simplicity bias is an important source of poor OOD generalization (studied e.g. with MNIST/CIFAR collages in~\cite{shah2020pitfalls}).
Our study is however not strictly tied to the simplicity bias: the proposed method improves the sampling of the hypothesis space, regardless of the default solution being explained by the simplicity bias or by other effects, such as neural anisotropies for example~\cite{ortiz2021neural}.

%\myparagraph{Q11: \textit{``''}}
%We define ``side knowledge'' as task-specific information besides a set of labeled i.i.d. examples. The demarcation is not completely arbitrary: the latter alone is insufficient to specify a desired OOD behaviour.
%We did not claim that this should \textit{never} be used for training/designing architectures (e.g. translation or rotation invariance). Rather, we showed that \textit{some} of it could be used during model selection.

\myparagraph{Q10: \textit{``Can the method apply to complete ResNet-scale models?''}}
The extension to full models involves no conceptual leap since the regularizer operates in the space of representations (input or latent), not of weights.
The applicability of our findings to simple models with pretrained features is certainly useful in itself.
We haven't addressed the tuning of full-scale models so far due to our limitations in computational resources.

\myparagraph{Q11: \textit{``By relying on model selection to select the desired inductive biases, could the method be at greater risk of adaptive overfitting, since it requires more evaluations on the test set than other OOD approaches?''}}
Model selection is not limited to cross-validation, other choices include 
contrast sets~\cite{gardner2020evaluating}, calibration-based methods~\cite{wald2021calibration}, model explainability with expert knowledge, etc.
If cross-validation is used in a real application, OOD \textbf{validation} data would be used, never the test set data itself obviously.
%Correct. However, adaptive overfitting from cross-validation with OOD (``target'') data is well studied within the framework of classical learning theory (in-domain generalization) thus besides the point of this paper.
Our results under ``oracle selection'' serve to provide an upper bound on achievable OOD performance (i.e. with perfect model selection).
%without confounding it with ID generalization.
% (which is largely independent from our method and rather a function of the chosen hypothesis space/architecture).

\myparagraph{Q12: Clarification of some definitions.}
\textbf{Spurious correlations}: statistical pattern in the data that does not reflect a causal relationship (\ie not guaranteed to transfer OOD) but which results from confounding (\eg selection bias).
\textbf{Inductive biases}: assumptions made by a learning algorithm about the nature of the target function to enable finite-sample generalization.

%====================================================================================================================
\section{OOD Generalization and causality}
\label{secCausality}

The ultimate goal of supervised machine learning is to learn a model that mimics a real-world process that jointly determines the values of an observed variable $X$ and target variable $Y$.
It makes sense to learn a task when the process relates these variables with causal ($X\rightarrow Y$) mechanisms, such that the conditional $P(Y|X)$ is a property of the task: it will always remain the same across training and (OOD) test conditions.
%or anti-causal ($X\leftarrow Y$) 
The OOD setting used throughout this paper is also called \emph{covariate shift}: $P(Y|X)$ is constant and only the marginal $P(X)$ varies across training and test sets.
For example, $X$ can represent images being drawn as photographs during training and paintings during testing.

To achieve OOD generalization across arbitrary covariate shifts, it is necessary that the causal mechanisms of inference (within the learned model) mirror the causal mechanisms of the data-generating process.
In other words, features responsible for the prediction of a label by the model should be the same as those causally related to this label in the training data.
However, causal properties of the real-world process are \emph{not} properties a joint distribution over $(X,Y)$ of training examples~\cite{scholkopf2021toward}.
\textbf{The information necessary for OOD generalization is lost} by drawing i.i.d. training samples from the data-generating process.
This is why optimizing a model for ERM cannot generally achieve the above condition.
\textbf{Even infinite amounts of training data} cannot bring back this missing information.

Note that all of the above is true for arbitrary tasks and data distributions. If task- or dataset-specific knowledge is available, it sometimes sufficiently circumscribes the space of possible ground-truth data-generating processes
to allow recovering some causal properties from observational data (see examples with images~\cite{lopez2017discovering} and time series~\cite{klindt2020towards}).
The method of this paper makes use of no such prior knowledge.

%====================================================================================================================
\section{Project chronology and negative results}

This section is an informal chronology of the developments that led to this paper, including things we tried that did not work.
\vspace{5pt}

\myparagraph{Initial motivation.}
The initial motivation for this work was to learn patterns that a model would not learn by default because of the simplicity bias.
The closest existing works were ``debiasing methods'' popular for visual question answering~\cite{cadene2019rubi,clark2019don,jing2020overcoming} and NLP~\cite{bahng2020learning,clark2020learning,mahabadi2019simple,stacey2020avoiding,utama2020mind}.
They typically train one model that is biased by design (for example being fed a partial input) while a second model is subsequently trained to be different, hence more robust.
Our first innovation was to enforce this difference in the space of input gradients of these models (rather than in the space of their activations, weights, etc.).
This quickly appeared effective and easier to train than adversarial objectives of many debiasing methods.
A later literature search showed that input gradients had previously been used in similar~\cite{kariyappa2019improving} and other applications~\cite{du2018adapting}.

%We worked with toy data.
To improve over existing debiasing methods, we extended our approach to ${>}2$ models and removed the reliance on a ``weak'' first model by design.
Instead, we used the same architecture for all models and realized the simplicity bias would make any model ``weak'' by default.
The publication of multiple studies in late 2020 related to the simplicity bias encouraged pursuing with this approach.

\myparagraph{Parallel training.}
Our first implementation used sequential training of multiple models, inspired by existing debiasing methods.
We implemented a parallel version as a baseline, but it quickly proved more effective, to our surprise.
Despite much effort, the sequential scheme could not equate the parallel version.
We gave up the former after coming up with a satisfying intuitive explanation.
The sequential training makes each model only marginally different from the previous one (for example, with the collages, every model would use another pixel of the MNIST digit, but would never focus on a completely different part of the images).
This holds even after training a very large number of models, and whether the diversity regularizer is applied on the last two models, or on the whole collection of models trained so far.
In contrast, the parallel training, using with pairwise constraints between all models simultaneously, could produce multiple potentially-good models at once.
%, and without excessive tweaking of hyperparameters.

\myparagraph{Similarity of gradients.}
To achieve the goal of maximizing the diversity of the models, we designed many elaborate measures of similarity between the gradients.
The intuitive goal was to ``spread'' the learned solutions evenly within the space of predictors.
However, none of these alternative measures worked better than the sum of pairwise dot products described in this paper.
Alternatives that we tested include cosine distances, the determinant of a matrix of dot products (similar to determinantal point processes or DPPs) or of other kernel function of the dot products, a soft approximation (logSumExp) of the maximum of the pairwise dot products (rather than the overall sum).

We also tried using the gradient w.r.t. the spatial input (all input pixels), or w.r.t. intermediate representations in CNNs.
We also tried all of these as gradient-feature products (in the style of the Grad-CAM method).
The bare gradients worked great with the collages, and the gradient-feature products worked clearly better with ResNet as the feature extractor. 
Our larger-scale experiments therefore all use the gradient-feature products.

We also tried applying various normalizations (L0, L1, L2, softmax: none worked) and rectifications to the gradients (absolute value, ReLU/positive part, negative part, square: the square did not work at all, but all other options performed similarly).

As described in the paper, we use the gradient of the top predicted score. Using the gradient of its corresponding logit (before sigmoid rectification) seems to work equally well.
We tried alternatives: the gradient of the score predicted for the ground truth class, or the gradient of the classification loss. Both performed worse.

\myparagraph{Datasets.}
We started with toy data (32-dimensional vectors of numeric values generated with known functions) in the style used in~\cite{parascandolo2020learning}.
We then moved to colored MNIST digits. This dataset proved useless since the signal is perfectly disentangled across two input channels, which is ridiculously simple and unrealistic.
We found the multi-dataset collages~\cite{shah2020pitfalls} the best compromise between toy and real data.
We then moved to the real datasets PACS and BAR.
Overall, most of our developments were done with the collages and PACS (using \emph{art\_painting} as the test style because it seemed to be the PACS setting with the clearest possible improvements, from results of other methods).

%====================================================================================================================
\section{Related work}
\label{appendix-relatedWork}

We provide below an extended literature review.
Since this paper addresses a central problem in machine learning, it touches many well-established research areas.

%====================================================================================================================
\paragraph{Importance of OOD generalization.}
Failure to generalize OOD is the root cause of many limitations of machine learning: adversarial attacks~\cite{hendrycks2019benchmarking}, some model biases~\cite{nanda2020fairness}, failure to generalize across datasets~\cite{torralba2011unbiased}, etc.
%These symptoms have cause an increasing awareness of the limitations of models trained with ERM.
% with many examples in computer vision~\cite{??} and natural language processing (NLP)~\cite{??}.
%These limitations often become apparent only when testing models on OOD data.
Poor OOD generalization is only apparent and problematic with OOD test data.
Academic benchmarks have traditionally been built with i.i.d. training and test samples.
This rarely holds in the real world, and OOD is closer to the norm in real-life deployments of machine learning models.

Evaluation with i.i.d. training/test sets hide a model's limitations because spurious correlations and biases in the training data also exist in the test data.
Thanks to the increasing awareness of issues of robustness with deep learning, many benchmarks now include OOD (a.k.a. ``challenge'') tests sets~\cite{gardner2020evaluating,gorman2019we}.
OOD evaluation can also be done by cross-dataset evaluation without fine-tuning (\ie zero-shot transfer)~\cite{radford2021learning}.
The assumptions then is that the spurious patterns in different datasets are uncorrelated.
%The result is an inflated sense of progress for the research community.
%This has plagued a number of benchmarks in computer vision and natural language processing, sometimes going unnoticed for years~\cite{nlu see email with peter anderson,summarization,vqa}.

% references adv regularization:
% Explaining and harnessing adversarial examples
% Towards deep learning models resistant to   adversarial attacks
% A closer look at accuracy vs. robustness

%====================================================================================================================
\paragraph{Improving OOD generalization.}
Given its central place, OOD generalization is addressed from multiple angles by multiple communities making different assumptions.
\textbf{Domain generalization} methods~\cite{gulrajani2020search} use multiple domains during training.
\textbf{Domain adaptation} methods use unlabeled data from a second domain for rapid adaptation at test time.
\textbf{Debiasing} methods use expert knowledge of the spurious correlations to prevent the model from using them.
%by controlling inductive biases (with architectures, regularizers).
\textbf{Adversarial training} methods use expert knowledge of the type of statistical patterns that are undesirable to learn, or notions of smoothness and continuity that a model should exhibit.
%, also controlling inductive biases through various ways.
Other training objectives have also been proposed to use \textbf{interventional data} such as counterfactual examples~\cite{heinze2017conditional,teney2020learning} or non-i.i.d. datasets like non-stationary time series~\cite{halva2020hidden,hyvarinen2017nonlinear,pfister2019invariant}.
%The success of data augmentation was also recently explained as constituting one such kind of interventional data~\cite{??} using the theory of causal reasoning to support a long history of empirical results.
The common point to all approaches that improve OOD generalization is that extra knowledge is provided, either as expert task-specific knowledge, or as non-i.i.d. data.
This corroborates the point made throughout this paper that i.i.d. data alone (even in infinite quantity) cannot improve OOD generalization.

%====================================================================================================================
\paragraph{Domain generalization.}
The goal of domain generalization (DG) is to learn models that generalize across visual domains such as photographs, sketches, paintings, etc.
Images from multiple domains (a.k.a. training environments) are provided for training. The model is then evaluated on one held-out domain.
The training environments can be formalized as different interventions on the data-generating process.
This was shown to carry the kind of information required for OOD generalization~\cite{arjovsky2019invariant,peters2016causal}.
Intuitively, DG methods discover features of the input that are ``common'' and similarly-predictive across the environments.
%, assuming they can also be relied on on the unseen test domain.
The principle is sound if a large number of training domains is available but this is not the case with existing datasets. PACS for example provides only three training domains.
Although some of the information necessary for OOD generalization is theoretically there, the learning problem is still very ill-defined because of this large distribution shift between the training domains.
Practically, this leaves much room to apply various inductive biases. This explains the plethora of methods already developed for this dataset.
Because the problem is ill-defined, the effectiveness of any such method can only be assessed when confronted with the test domain (and this is what we also need to do after training a collection of models with our method).

Gulrajani and Lopez-Paz~\cite{gulrajani2020search} discussed the practice of model selection using the test domain.
They observed that no existing method performed better than ERM \textbf{when access to the test domain is restricted}.
This is unsurprising to us: this follows from the fundamental need, to achieve generalization, of substantial knowledge about the relation between training and test distributions.
And this information is unlikely to be available from a handful of disparate training domains.

%The most popular strategy is to learn useful features while constraining them to have similar distributions across domains (Ganin et al., 2016; Li et al., 2018b).
%Other works constrain these features in a way that one can learn a classifier that is simultaneously optimal across all domains (Peters et al., 2016; Arjovsky et al., 2019; Krueger et al., 2020).
%meta-learning (Li et al., 2018a), parameter-sharing (Sagawa et al., 2020a) and data augmentation (Zhang et al., 2018).

%It is hardly surprising that limiting the model selection in~\cite{gulrajani2020search} (limited number of tested hyperparameters, no early stopping) does not yield models superior to ERM.
%the central point, which is that model selection in domain generalization is nontrivial because we do not know any information about the test distribution at training time.

%The baseline ERM approach is to mix all training examples together.
%Advanced methods use the labels of the training domains to discover which features hold across domain, and are thus likely to also be relevant at test time.

Our method does no require the labels of training environments used by DG methods.
Our setup is more similar ``single-source'' domain adaptation~\cite{volpi2018generalizing,qiao2020learning}.
These methods augment the training data using a generative model to expand the region of feature space in which the predictions of the model are ``stable''.
Consequently, these methods cannot make the model use features of the data it was not using in the first place. Thus there is no hope to counter the simplicity bias.

\paragraph{Debiasing.}
Methods for debiasing are concerned with improving generalization of models against a precisely identified (undesirable) factor of variation in the data~\cite{bahng2020learning,clark2020learning,mahabadi2019simple,nam2020learning,stacey2020avoiding,utama2020mind}.
In computer vision for example, this can be removing the bias towards texture in the ImageNet dataset~\cite{baker2018deep,geirhos2018imagenet}.
In NLP, this can be removing the ``hypothesis-only bias'' of entailment models, that make these models guess an answer without considering the whole input~\cite{clark2020learning}.

Debiasing is relevant to this paper because most methods rely on training multiple models that each use the input differently.
The source of improvement is the explicit specification of the factor of variation to be be invariant to.
Typically, debiasing methods train a pair of models to respectively focus or ignore it. The latter model is used at test time.
For example in~\cite{bahng2020learning}, a first CNN model is trained with an architecture providing a small receptive field, such that is focuses on local texture.
A second CNN is then trained with a larger receptive field while a regularizer makes its activations uncorrelated with the first one's, such that it focuses on overall shape more than on texture.
Variations of the method include the extension to more than two models~\cite{stacey2020avoiding}.
%	Avoiding the Hypothesis-Only Bias in Natural Language Inference via Ensemble Adversarial Training
%	Adversarial removal of demographic attributes from text data
%	End-to-end bias mitigation by modeling biases in corpora
%These methods rely on the explicit specification of one factor of variation that the model should ignore.
In comparison to our work, debiasing methods require the explicit specification of a factor of variation to ignore and they require it to be easily disentangled from other features of the input.

Some debiasing methods claim to require \textbf{no explicit knowledge of the bias}~\cite{utama2020towards,clark2020learning,sanh2020learning}.
They actually make this knowledge only less explicit: the authors design the architecture of the models such that the weak learner is forced to use the bias (through limited capacity, receptive field, etc.).
%The strong model then knows what to ignore.
Our method does not rely on heterogeneous architectures and is applied to many more models.
We also found that parallel training of multiple models was much more effective than the sequential training used with most debiasing methods.
%We also do not assume that the ``strong'' learner will be best, since our model selection can still select a simple model from all trained candidates.

%====================================================================================================================
\paragraph{Encouraging diversity within one model.}

We can draw parallels between the method in this paper and existing methods that encourage a notion of diversity.
Classical feature selection approaches~\cite{guyon2003introduction} are related but they not suitable to deep learning model and high-dimensional representations.
For example, SCOPS~\cite{hung2019scops} performs self-supervised part discovery using an objective of orthogonality (akin to diversity) between parts.
In comparison, we use diversity as an objective alongside predictive performance.
Diversity was also used an objective during model compression, for fusing redundant neurons with similar activations~\cite{mariet2015diversity}.
%Diversity Networks: Neural Network Compression Using Determinantal Point Processes
Closer technically to our method,~\cite{du2018adapting} uses the cosine similarity of gradients of multiple losses to measure their mutual correlation in the context of multitask learning.
Previous works~\cite{kariyappa2019improving,pace2020learning} proposed to promote diversity in the space of learned representations within a model.
Our approach is different in that we promote diversity across multiple independent models.
These works focus on synthetic data and adversarial robustness while we show improvements on multiple benchmarks with real data.

%The method in this paper encourages diversity in a collection of models by minimizing pairwise dot products between gradient-based descriptions of these models.
%This was used in the context of improving the usefulness of auxiliary tasks to improve data efficiency.
%Pace et al. adversarial training = min-max problem, difficult to train; only toy data 

%====================================================================================================================
\paragraph{Encouraging diversity within ensembles.}

Ensembling several models is a common technique to improve predictive performance over a single model.
The diversity of the models in an ensemble is important~\cite{yu2011diversity} and usually promoted by training models with different hyperparameters or random seeds, or by enforcing diversity in the space of weights of the models~\cite{xie2015generalization}.
In comparison, our diversity loss operates in the space of the gradients of the models. They need to be differentiable w.r.t. their input but their implementation as neural networks is irrelevant.
For \textbf{adversarial robustness}, ensembles have shown benefits.
Both~\cite{bagnall2017training} and~\cite{pang2019improving} encourage diversity in the distributions of the models' logits.
\cite{kariyappa2019improving} minimizes the cosine similarity between gradients of the models.
For \textbf{domain generalization},~\cite{chattopadhyay2020learning} learns an ensemble of classifiers on CNN features.
Each classifier is trained on a different visual domain and they promote diversity by minimizing the overlap between features used by each model, such that each specialize to one domain.
% Learning to Balance Specificity and Invariance for In and Out of Domain Generalization

The multiverse loss~\cite{littwin2016multiverse,malkiel2021maximal} improves transfer learning by duplicating a cross-entropy loss over multiple linear classifiers with an orthogonality constraint on their weights.
It was shown to increase the number of distinct discriminative directions of the learned representation.
It can be seen as a special case of our method.

In~\cite{ross2017right}, the authors use input gradients to sequentially train multiple copies of a model to focus on different input features. The authors however mentioned in private communication that ``sequential training doesn't work in most cases'' which has also been our experience (see our negative results in the appendix). Their experiments are limited to toy datasets.
After the writing of this paper, we realize that the same authors subsequently honed in on a parallel training scheme and diversity regularizer very similar to ours~\cite{ross2020ensembles,ross2018learning}, using a cosine similarity of gradients. Our motivation and experiments are very different and we invite the reader to consult those papers for a complementary view.

In~\cite{ratzlaff2019hypergan}, the authors train a generative model (HyperGAN) of parameters for a chosen network architecture. Their goal is to produce a diverse set of models, which they enforce and evaluate in parameter space. We think that our use of input gradients is more implementation-invariant and better capture the overall function implemented by deep models. We also believe that our evaluation with OOD tasks better captures the functional diversity of the learned models.

The earliest use of input gradients of neural networks was proposed by Drucker and LeCun~\cite{drucker1992improving} as ``double backpropagation'' to improve in-domain generalization. Almost identical formulations were described in many subsequent papers; see~\cite{hoffman2019robust} and citations therein.

\section{Experimental details}
\label{appImplementation}

%====================================================================================================================
\myparagraphWoSpacing{Collages dataset.}
\label{appCollagesDataset}
We use images from MNIST, Fashion-MNIST, CIFAR-10, SVHN.
The images are converted to grayscale. The images from MNIST and Fashion-MNIST are padded to $32{\times}32$ pixels.
We pre-select two classes from each dataset to be respectively associated with the collages $0$ and $1$ labels.
We follow~\cite{shah2020pitfalls} and choose 0/1 for MNIST, automobile/truck for CIFAR-10, and additionally choose 0/1 for SVHN and pullover/coat for Fashion-MNIST.
We generate a training set of 51,200 collages, and multiple test sets of 10,240 collages each.
Each collage is formed by tiling four blocks, each containing an image chosen at random from the corresponding source dataset.
The images in our training/evaluation sets are selected respectively from the original training/test sets of the source datasets.

We propose two versions of the dataset. An \emph{ordered} version, where the four blocks appear in constant order, and a \emph{shuffled} version where the order is randomized in every collage.
The shuffled version can be used to demonstrate that a given method does not rely on a known or constant image structure.

In the training set, the class in each block is perfectly correlated with collage label.
In each of the four test sets, the class in only one block is correlated with the collage label. The other blocks are randomized to either of its two possible classes.
We also generate four training sets in this manner, used solely to obtain upper bounds on the highest accuracy achievable on each block with a chosen architecture.

\myparagraph{Collages experiments.}
\label{appCollagesModels}
We use the \emph{ordered} version of the dataset. This allows generating the visualizations of Figure~\ref{figGrads} and the use of a simple fully-connected classifier.
We initially downsample the images by a factor $4$.
In our models, the feature extractor is the identity function and the classifier is a fully-connected MLP with two hidden layers of size $16$ with leaky ReLU activations (leak rate: $0.01$).
The classifier is followed by a sigmoid and trained to minimize a standard binary cross-entropy loss.
Training is performed by SGD with Adam, a learning rate of $0.001$, mini batches of size $256$, for a fixed number of $65$ epochs ($13$k iterations) with no early stopping.
The diversity regularizer is implemented as described in the paper.
The visualizations in Figure~\ref{figGrads} are obtained with the same model trained on images downsampled by a factor $16$.
The input gradient is evaluated and averaged on randomly selected test images.
They are then upsampled by bilinear interpolation to the original collage dimensions of $64{\times}64$ for visualization purposes (without upsampling, they obviously look more ``blocky'').

%====================================================================================================================
\myparagraph{BAR Experiments.}
\label{appBar}
We follow~\cite{nam2020learning} and use frozen features from a standard pretrained ResNet-50.
We train a 2-hidden-layer MLP classifier on these features.
We spent some effort optimizing this baseline to the point of almost equating the method proposed in~\cite{nam2020learning} (they only used a linear classifier on ResNet features).
We use Adam, a learning rate of $0.001$, a batch size of $256$, and hidden layer dimensions of $512$.
We applied our method on this strong baseline.

As usually done with ResNet, feature maps are globally pooled before the classifier.
Thus, unlike the experiments on collages, the features $\bh$ have no spatial dimensions.
The input gradients are compared across channels instead.
We implement this with a variant of Eq.~(\ref{eqDotProd}) inspired by the Grad-CAM method~\cite{selvaraju2017grad}, in which we multiply the gradients (with respect to the features) by the features themselves:
\abovedisplayskip=6pt
\belowdisplayskip=5pt
\begin{equation}
  \delta_{g_{\bphi_1},g_{\bphi_2}}(\bh) ~=~
  (\bh \, \nabla_\bh g^\star_{\bphi_1}) . (\bh \, \nabla_\bh g^\star_{\bphi_2})
  \,.
  \label{eqDotProd2}
\end{equation}

Each model is trained for 200 iterations with no early stopping.
The optimal weight of the regularizer is found by selection on the OOD test set.
For the reasons explained at length in the paper, we found no reliable strategy to tune it without access to the OOD test set.
This strategy is used for both our method and all other regularizers, ensuring a fair comparison.

%====================================================================================================================
\myparagraph{PACS Experiments.}
\label{appPacs}
We use a standard ResNet-18 as the feature extractor like most current methods.
\cite{gulrajani2020search} showed that a ResNet-50 could slightly improve performance but we did not have the computational resources to do so, and the results of most methods to compare ours with also use ResNet-18.
We first fine-tuned a ResNet-18 in the standard ``ERM'' setup (aggregated data of three training domains, linear output layer on top of ResNet features, Adam optimizer, learning rate of $4e{-}5$, batch size of 32, with augmentations described in~\cite{gulrajani2020search}, and early stopping based on test set accuracy).
We found that training with a sigmoid activation and binary cross-entropy loss was slightly better than the usual softmax. All our PACS models (baselines and others) therefore use a sigmoid output.
All of these choices provided a very strong baseline on which to test our method. Our baseline is noticeably stronger than those in existing papers as noted in Table~\ref{tabPacsAllStyles}.
Demonstrating an improvement over a strong baseline is obviously more challenging than over a weaker one.

Our method was with a two-hidden-layer MLP as the classifier, fed with frozen features from the fine-tuned ResNet-18 (hidden size of 512, leaky ReLUs of rate $0.01$, Adam, learning rate $3e{-}5$, batch size $256$).
We use the input gradient over channels (not over spatial dimensions) described above (Eq.~\ref{eqDotProd2}).

%All experiments in this paper were run on a single laptop with a GeForce GTX 1050 Ti GPU.
%All code was implemented in Matlab and is available at \url{http://<anonymized>}. The version R2021a or newer is required to compute second derivatives and enable training with our gradient-based regularizer.
%Unlike existing works we refrain from reporting results with 2 decimal digits. This would artificially give a sense of accuracy whereas the noise in the evaluation from stochastic factors and minute hyperparameters is much larger than that.
%We rather focus on the qualitative improvements highlighted in our ablative experiments.

%====================================================================================================================
\myparagraph{Existing regularizers.}
\label{appBaselines}
We describe below the existing regularizers reported in Tables~\ref{tabCollagesShort},~\ref{tabBar}, and~\ref{tabPacsAblation}.
\setlist{nolistsep,leftmargin=*}
\begin{enumerate}[topsep=1pt,itemsep=2pt]
  \item \textbf{Dropout.} Used on collages only. Dropout is applied on input images \ie on pixels of the quarter-size images fed to the MLP. We tuned the dropout rate, hoping that very high dropout rates would force the model to learn different parts of the image, but it did not work.
  \item \textbf{Penalty on L1 norm of gradients.} This adds the following term to the minimization objective: $|| \nabla_\bh g ||_1$.
  \item \textbf{Penalty on L1 norm of feature-gradient product.} Variant that uses the same product as our regularizer with BAR and PACS (Eq.~\ref{eqDotProd2}): $|| \bh \, \nabla_\bh g ||_1$.
  \item \textbf{Penalty on squared L2 norm of gradients.} Also known as Jacobian regularization~\cite{hoffman2019robust}, it adds the following term to the minimization objective: $|| \nabla_\bh g ||^2_2$.
  \item \textbf{Penalty on squared L2 norm of ReLU of gradient.} Variant that only uses the positive coordinates of the gradient:  $|| \operatorname{ReLU}(\nabla_\bh g) ||^2_2$. The rationale is that this variant is ``half-way'' like the feature-gradient product described below which also masks some coordinates of the gradient (the features come from a ReLU and have a number of coordinates equal to zero).
  \item \textbf{Penalty on squared L2 norm of feature-gradient product.} Variant that uses the same product as our regularizer with BAR and PACS (Eq.~\ref{eqDotProd2}): $|| \bh \, \nabla_\bh g ||^2_2$.
  \item \textbf{Penalty on squared L2 norm of logits.} Also known as spectral decoupling~\cite{pezeshki2020gradient}, it adds the following term to the minimization objective: $||  g ||_2^2$.
\end{enumerate}
All penalty terms are summed over all training examples.

\vspace{12pt}

\begin{table}[h!]
  \footnotesize
  \renewcommand{\tabcolsep}{0.3em}
  \renewcommand{\arraystretch}{1.0}
  \centering
  \begin{tabularx}{\linewidth}{Xccccc}
  \toprule
  Collages dataset & \multicolumn{4}{c}{Accuracy (\%) on} \vspace{1.5pt}\\ \cline{2-5}
  ~ &  \rotatebox{90}{MNIST} & \rotatebox{90}{SVHN~~} & \rotatebox{90}{Fashion-M.\phantom{e}} & \rotatebox{90}{CIFAR-10}\\
  \midrule
  \multicolumn{5}{l}{\textbf{Upper bounds}: training data with all blocks but one randomized}\vspace{4pt}\\
  MNIST      predictive only & \textbf{99.7} & 49.7 & 50.5 & 49.9 \vspace{-3pt}\\ ~ & \pp{0.0} & \pp{0.0} & \pp{0.0} & \pp{0.0} \\
  SVHN       predictive only & 50.2 & \textbf{89.8} & 50.5 & 50.2 \vspace{-3pt}\\ ~ & \pp{0.2} & \pp{0.5} & \pp{0.1} & \pp{0.2} \\
  Fashion-M. predictive only & 50.0 & 50.3 & \textbf{77.3} & 49.0 \vspace{-3pt}\\ ~ & \pp{0.2} & \pp{0.1} & \pp{0.4} & \pp{0.4} \\
  CIFAR      predictive only & 49.9 & 50.1 & 50.2 & \textbf{68.4} \vspace{-3pt}\\ ~ & \pp{0.2} & \pp{0.3} & \pp{0.4} & \pp{0.9} \\
  \midrule
  \multicolumn{5}{l}{\textbf{With proposed regularizer}, 32 models}\vspace{4pt}\\
  Best model on MNIST      & \textbf{95.4} & 49.6 & 50.6 & 49.8 \vspace{-3pt}\\ ~ & \pp{0.4} & \pp{0.1} & \pp{0.3} & \pp{0.1} \\
  Best model on SVHN       & 51.0 & \textbf{79.3} & 52.2 & 51.4 \vspace{-3pt}\\ ~ & \pp{2.7} & \pp{3.1} & \pp{1.2} & \pp{1.1} \\
  Best model on Fashion-M. & 50.6 & 50.4 & \textbf{69.0} & 52.6 \vspace{-3pt}\\ ~ & \pp{1.6} & \pp{0.3} & \pp{3.0} & \pp{1.6} \\
  Best model on CIFAR      & 50.4 & 50.3 & 56.1 & \textbf{59.6} \vspace{-3pt}\\ ~ & \pp{1.5} & \pp{0.3} & \pp{1.2} & \pp{0.5} \\
  \bottomrule
  \end{tabularx}
  \vspace{4pt}
  \normalsize
  \caption{%
    Detailed results on collages. We identify the best model on each test set as in Table~\ref{tabCollagesShort}. The difference is that we report the accuracy \textbf{on all four test sets} (in the main paper, these were summarized as a single row).
    Each model specializes and is good on a single test set at a time. This shows that the features used by each model do not overlap.
    \label{tabCollagesLong}}
  \vspace{-12pt}
\end{table}

\begin{figure*}[h]
	\centering
	\includegraphics[width=.30\linewidth]{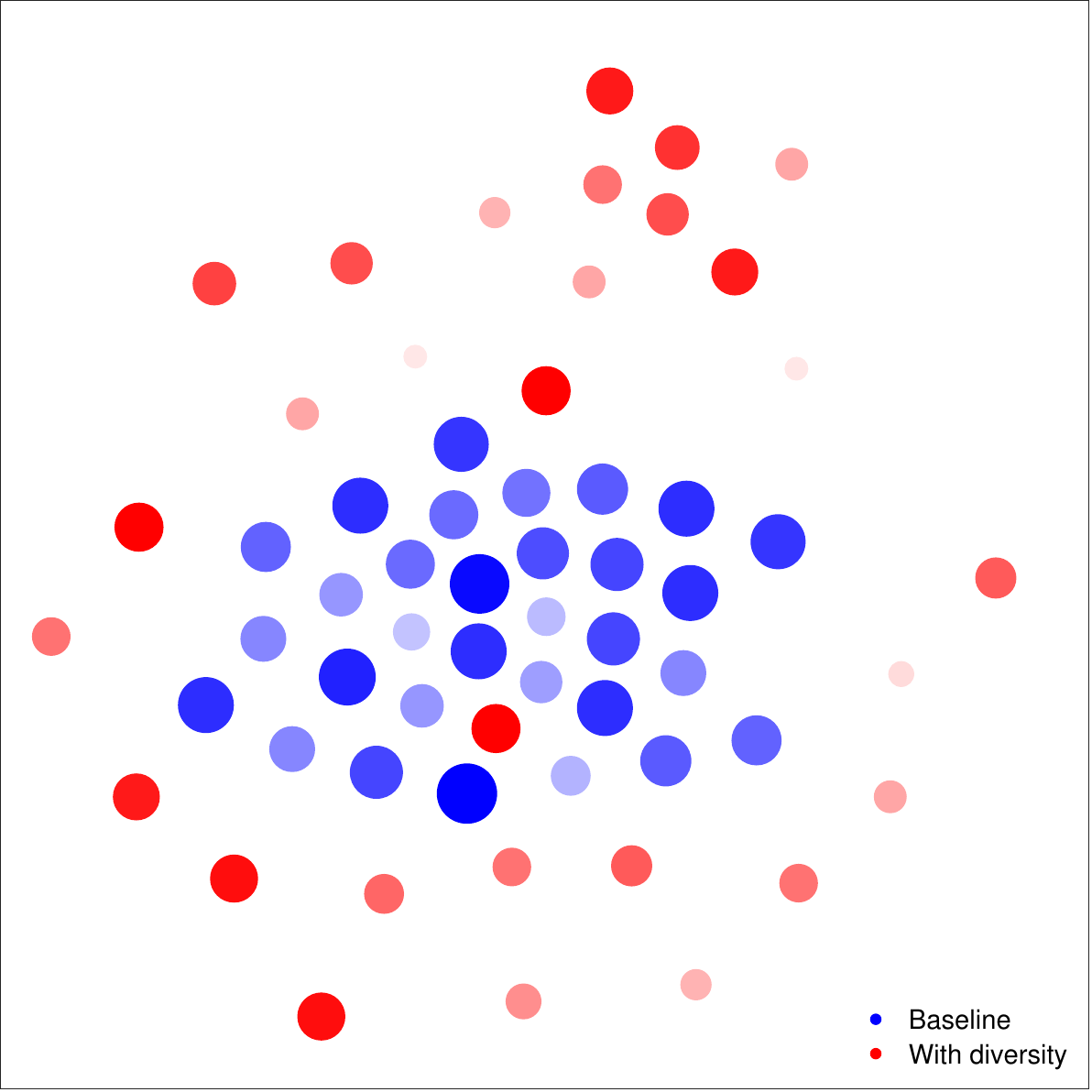}
	\vspace{2pt}
	\caption{%
  2D Projection of the gradients $\{ \nabla_\bh g^\star_{\bphi_i} \}_{i{=}1}^{64}$
  of a collection of models trained \textcolor{blue}{without} and \textcolor{red}{with} our diversity regularizer (PACS dataset, ``art'' as test style).
  Each point represents one model. With the regularizer (\textcolor{red}{in red}), the gradients are clearly more spread out.
  The 2D projection is done with t-SNE using the inverse of the dot product as the distance function. The size and color saturation of each point are proportional to the accuracy of the corresponding model.
  \label{figGradients}}
	%\vspace{-8pt}
\end{figure*}

\begin{figure*}[h ]
	\centering
  \footnotesize
  \renewcommand{\tabcolsep}{1em}
  \renewcommand{\arraystretch}{4.6}
  \begin{tabular}{rl}
    \hangBoxC{\begin{tabular}{r}
        \vspace{-65pt}\\Climbing\\Diving\\Fishing\\Racing\\Throwing\\Vaulting
    \end{tabular}}&
    \hangBoxC{\includegraphics[width=.85\linewidth]{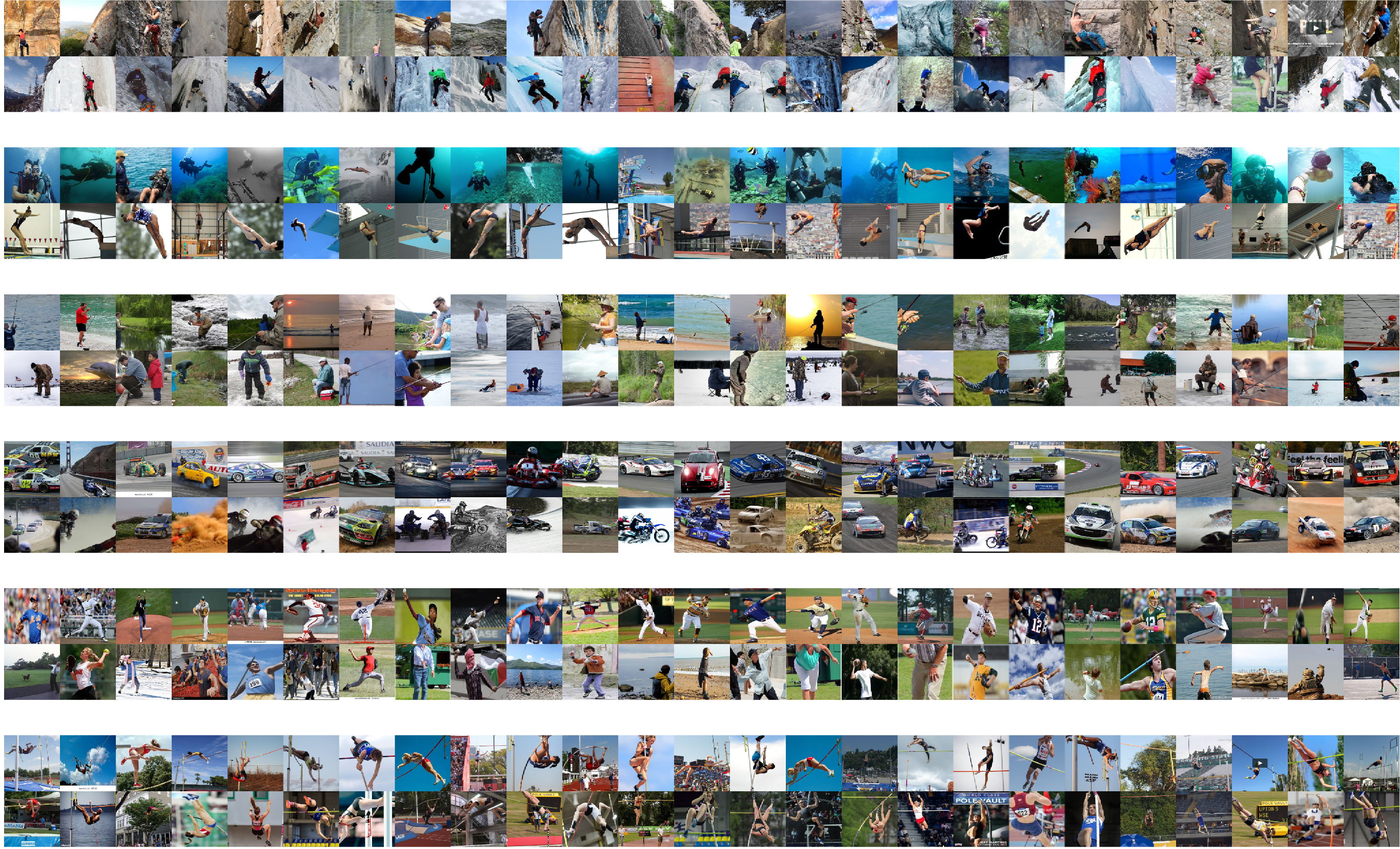}}\\
  \end{tabular}
	%\vspace{-6pt}
	\caption{Examples from the biased activity recognition (BAR) dataset~\cite{nam2020learning}. Each row shows a different class, and the upper/lower part of each row shows training/test images respectively (for example on the first row, rock climbing/ice climbing).\label{figBarExamples}}
	%\vspace{-8pt}
\end{figure*}

%====================================================================================================================
%\section{Complementary theoretical support}
%\label{proofs}
%\input{proofsV2}

%====================================================================================================================
%\bibliographystyle{ieee_fullname}
%\bibliography{Bibliography}

\end{document}